\pdfoutput=1

\documentclass[11pt]{article}

\usepackage[final]{acl}

\usepackage{times}
\usepackage{latexsym}
\usepackage{amsmath} 
\usepackage{booktabs}
\usepackage{subcaption}  
\usepackage{graphicx} 
\usepackage[T1]{fontenc}
\usepackage[utf8]{inputenc}
\usepackage{multirow}
\usepackage{array}

\usepackage{microtype}

\usepackage{inconsolata}

\usepackage{graphicx}

\usepackage{todonotes}

\title{NeLLCom-Lex: A Neural-agent Framework to Study the Interplay between Lexical Systems and Language Use} 

\author{
 \textbf{Yuqing Zhang\textsuperscript{1}},
 \textbf{Ecesu Ürker\textsuperscript{2}},
 \textbf{Tessa Verhoef\textsuperscript{3}}, 
 \textbf{Gemma Boleda\textsuperscript{2,4}}, 
 \textbf{Arianna Bisazza\textsuperscript{1}
 }
\\
\\
 \textsuperscript{1}Center for Language and Cognition, University of Groningen \\
 \textsuperscript{2}Department of Translation and Language Sciences, Universitat Pompeu Fabra \\
 \textsuperscript{3}Leiden Institute of Advanced Computer Science,  Leiden University \\
 \textsuperscript{4}Catalan Institution for Research and Advanced Studies (ICREA)
\\
  \texttt{\small{\{yuqing.zhang, a.bisazza\}@rug.nl,}} 
\texttt{\small\{{ecesu.urker, gemma.boleda\}@upf.edu,}}
  \texttt{\small{t.verhoef@liacs.leidenuniv.nl}} \\
}

\begin{document}
\maketitle
\begin{abstract}
Lexical semantic change has primarily been investigated with observational and experimental methods; however, observational methods (corpus analysis, distributional semantic modeling) cannot get at causal mechanisms, and experimental paradigms with humans are hard to apply to semantic change due to the extended diachronic processes involved.
This work introduces NeLLCom-Lex, a neural-agent framework designed to simulate semantic change by first grounding agents in a real lexical system (e.g.\ English) and then systematically manipulating their communicative needs.
Using a well-established color naming task, we simulate the evolution of a lexical system within a single generation, and study which factors lead agents to:
(i) develop human-like naming behavior and lexicons, and (ii) change their behavior and lexicons according to their communicative needs. 
Our experiments with different supervised and reinforcement learning pipelines
show that neural agents trained to `speak' an existing language 
can reproduce human-like patterns in color naming to a remarkable extent, supporting the further use of NeLLCom-Lex to elucidate the mechanisms of semantic change.
%

\end{abstract}


\section{Introduction}

A central goal in the study of language emergence and evolution is to understand how cognitive and communicative pressures shape linguistic systems \citep{smith2004evolution, smith2022language, christiansen2008language,regier2015word}.
More broadly, different scholarly traditions and fields of study have stressed the dynamic relationship between language use, on the one hand, and the linguistic system, on the other, be it at the synchronic~\cite{goldberg95,hawkins2004efficiency,bybee2010language} or diachronic level~\cite{campbell2013historical,labov2010principles,hopper2003grammaticalization}.

In particular, the referential communicative context has been found to shape language use across timescales, ranging from pragmatically-driven naming choices in synchronous communication \citep{sedivy1999achieving, sedivy2003pragmatic}
to the longer-term emergence of shared lexical conventions and to their semantic shift in response to changing communicative needs (see, e.g., artificial language learning studies in \citet{winters2015languages} and \citet{hawkins2018emerging}).

While the relationship is widely acknowledged, studying the implicated mechanisms across all of these timescales is challenging. 
In recent decades, a large body of work has therefore focused on agent-based computational simulations \citep{hurford1989biological, steels1997synthetic,de2006computer}
, receiving a renewed increased interest with the deployment of modern neural-network agents \citep{lazaridou2018emergence, lazaridou2020emergent, chaabouni2021communicating, kharitonov-etal-2019-egg, lian2023communication}. 
We follow this tradition, focusing on the lexicon, which is essential for enabling reference to entities in the world and therefore plays a key role in effective communication \citep{steels2005emergence,brochhagen2022languages}.

Computational modeling studies have used communication games to simulate the emergence of lexical systems in neural network-based agents \citep{monroe-etal-2017-colors, white2020learning, kobrock-etal-2024-context, kobrock2025agents}. 
These methods enable control over contextual variables and make it possible to simulate language change across different timescales.
Despite these advances, most emergent communication studies begin with blank-slate agents and focus on how contextual variability shapes \textit{novel} communication systems (e.g., \citealp{kobrock-etal-2024-context}).
By contrast, little work has focused on modeling how an \textit{existing} lexical system might adapt pragmatically to changing communicative needs, 
an approach that allows both more direct comparisons to human behavior and the possibility of simulating specific trajectories of language change.

In this paper, we start exploring precisely this approach by introducing the NeLLCom-Lex framework. We build on NeLLCom \citep{lian2023communication}, a flexible agent-communication framework that combines generic supervised learning (SL) and reinforcement learning (RL) objectives, making it well-suited for exploring how communication pressures shape language use and lexical adaptation across different communicative contexts.


As a first exploration of the dynamics between use and structure, we focus on dyadic interactions and lexical evolution across a single generation; as well as on a specific contextual factor, namely the differences in the granularity of the distinctions speakers need to make in their communicative environments.
What we are thus simulating is linguistic adaptation across a lifetime, brought upon by communicative needs regarding reference.
Through a number of experiments with a well-established color naming task \cite{monroe-etal-2017-colors} and various evaluation methodologies inspired in the study of human behavior \cite{gualdoni-boleda-2024-objects}, 
we show that our NeLLCom-Lex agents can reproduce human-like pragmatic naming behavior, while maintaining overall efficient lexical systems.
%
We also find agent naming behavior changes in response to different communication needs in the environment.
Overall, our results support the use of NeLLCom-Lex to investigate the interplay between language use and language structure in the lexicon.%
\footnote{Code and materials are released at \url{https://github.com/yuqing0304/NeLLCom-Lex}.}

\section{Background}





It is known that, synchronically, speakers choose words that best identify the referent given the communicative context. 
For example, if a target object, e.g., a Dalmatian, appears in a context surrounded by other dogs, the word \textit{dog} may be too broad to identify the referent. In such cases, speakers tend to choose more specific words (e.g., \textit{Dalmatian}), reflecting a pragmatic naming choice driven by communicative need \citep{winters2015languages, graf2016animal}.
While it is easy to justify this kind of pragmatic adaptation to context, it is not so straightforward to explain why there is redundancy in the lexicon in the first place, in that it includes words at higher and lower levels of informativeness; why do languages not have only more informative words (\textit{Dalmatian}, \textit{Golden retriever}, \textit{chihuahua}), or why do they not only have less informative words that they modify (\textit{dog with black spots})?
\citet{gualdoni-boleda-2024-objects} hypothesized that this redundancy in the lexicon, coupled with pragmatic adaptation to context, is a good solution to the trade-off between communicative pressures to be understood and cognitive pressures to expend low effort.



Trade-offs between effort and informativeness in language use have been studied within the Rational Speech Acts (RSA) framework \citep{goodman2016pragmatic, graf2016animal, monroe-etal-2017-colors, white2020learning, zarriess-schlangen-2019-know}.
For instance, \citet{monroe-etal-2017-colors} collected experimental data from color reference games and built pragmatic neural speaker and listener models within the RSA framework. Experiments show that the embedded speaker model shows pragmatic behaviors, i.e., modulating utterance lengths based on contextual difficulty. 
Subsequent work by \citet{white2020learning} extended this line 
and found that only models explicitly trained through a pragmatic reasoning objective (or approximations thereof) were able to adjust utterance length to communicative context. 
In both of these studies, messages could be multi-word utterances, and utterance length was used as a measure of effort while variations in informativeness of individual word choices were not considered. 

Complementary research on emergent communication using RL highlights similar efficiency trade-offs, though primarily at the level of the lexical system \citep{lazaridou2018emergence, white2020learning, chaabouni2020compositionality, chaabouni2021communicating, kobrock-etal-2024-context, gualdoni2024bridging}. For example, replicating earlier findings \citep{zaslavsky2018efficient} from large-scale data analysis, \citet{chaabouni2021communicating} examined pressures in emergent communication among neural agents, focusing on the trade-off between communicative accuracy and complexity in color partitioning in agents' naming systems. Similarly, \citet{carlsson2024cultural} modeled combined pressures from communication and transmission of signals over generations 
to explain the emergence of efficient color naming systems. 
Overall, these studies focus on how general pressures coming from language learning and use shape properties of the lexical system, rather than on how pragmatic language use \emph{in context} shapes those properties.
While \citet{kobrock-etal-2024-context} studied emerging lexicons in context, 
their models developed languages from scratch using symbolic data, without grounding in real-world lexical meanings.

\begin{figure*}[t]
  \centering

  \begin{subfigure}{0.21\textwidth}
    \centering
    \includegraphics[width=\linewidth]{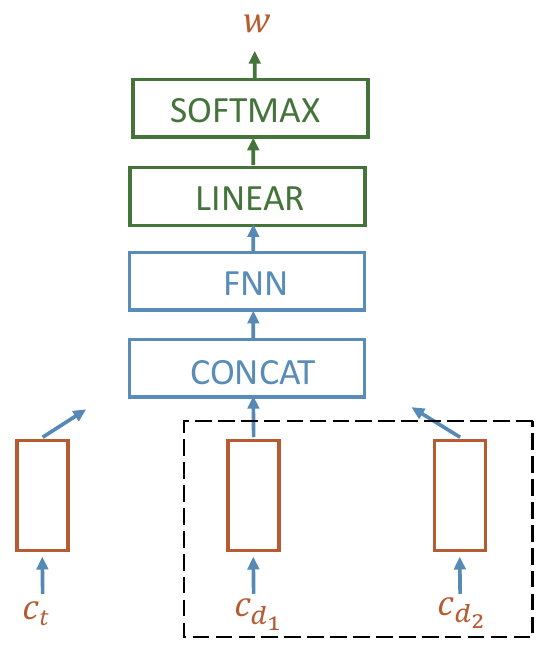}
    \caption{Speaker architecture}
  \end{subfigure}%
  \hfill
  \begin{subfigure}{0.21\textwidth}
    \centering
    \includegraphics[width=\linewidth]{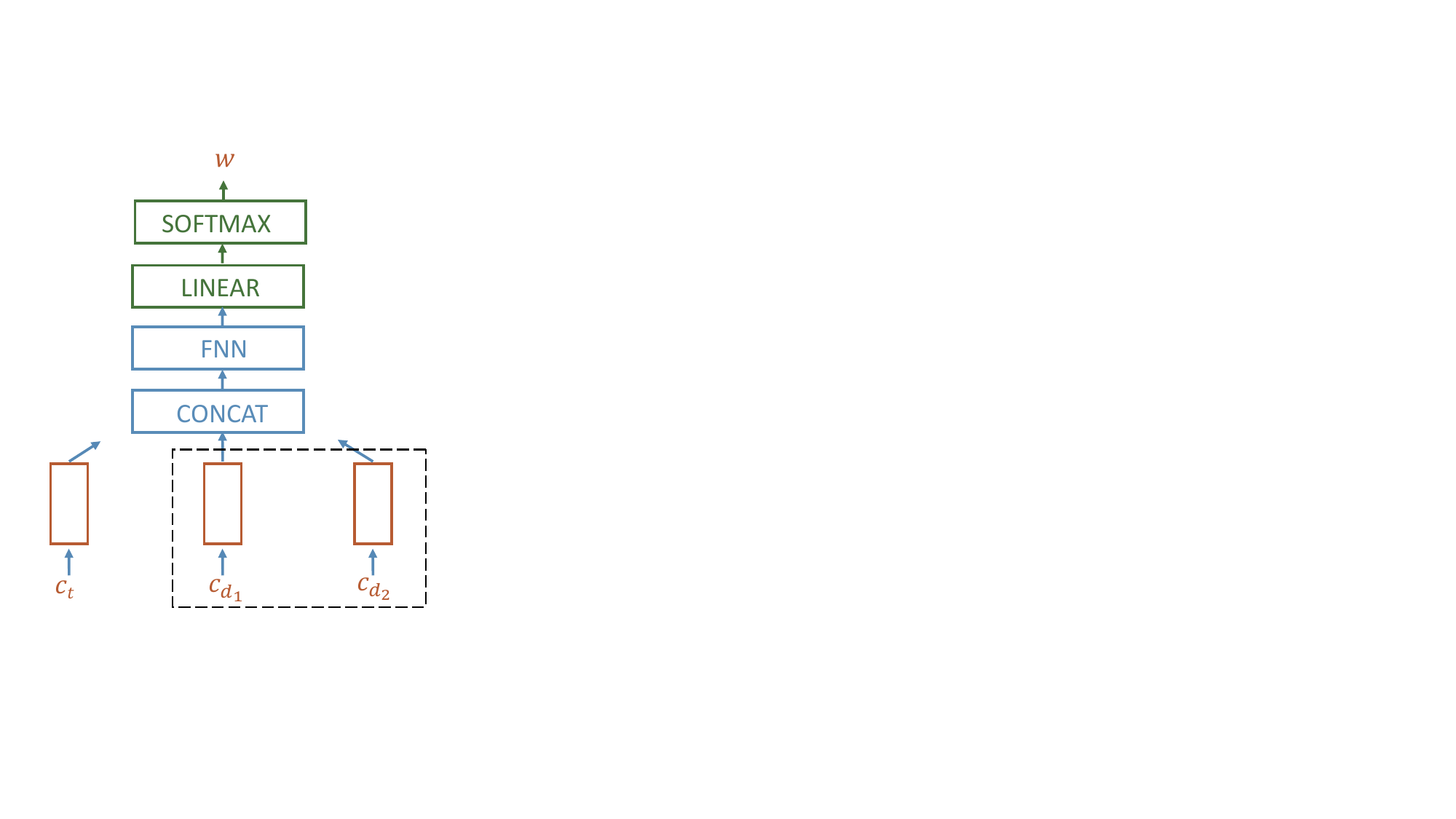}
    \caption{Listener architecture}
  \end{subfigure}%
  \hfill
  \begin{subfigure}{0.44\textwidth}
    \centering
    \includegraphics[width=\linewidth]{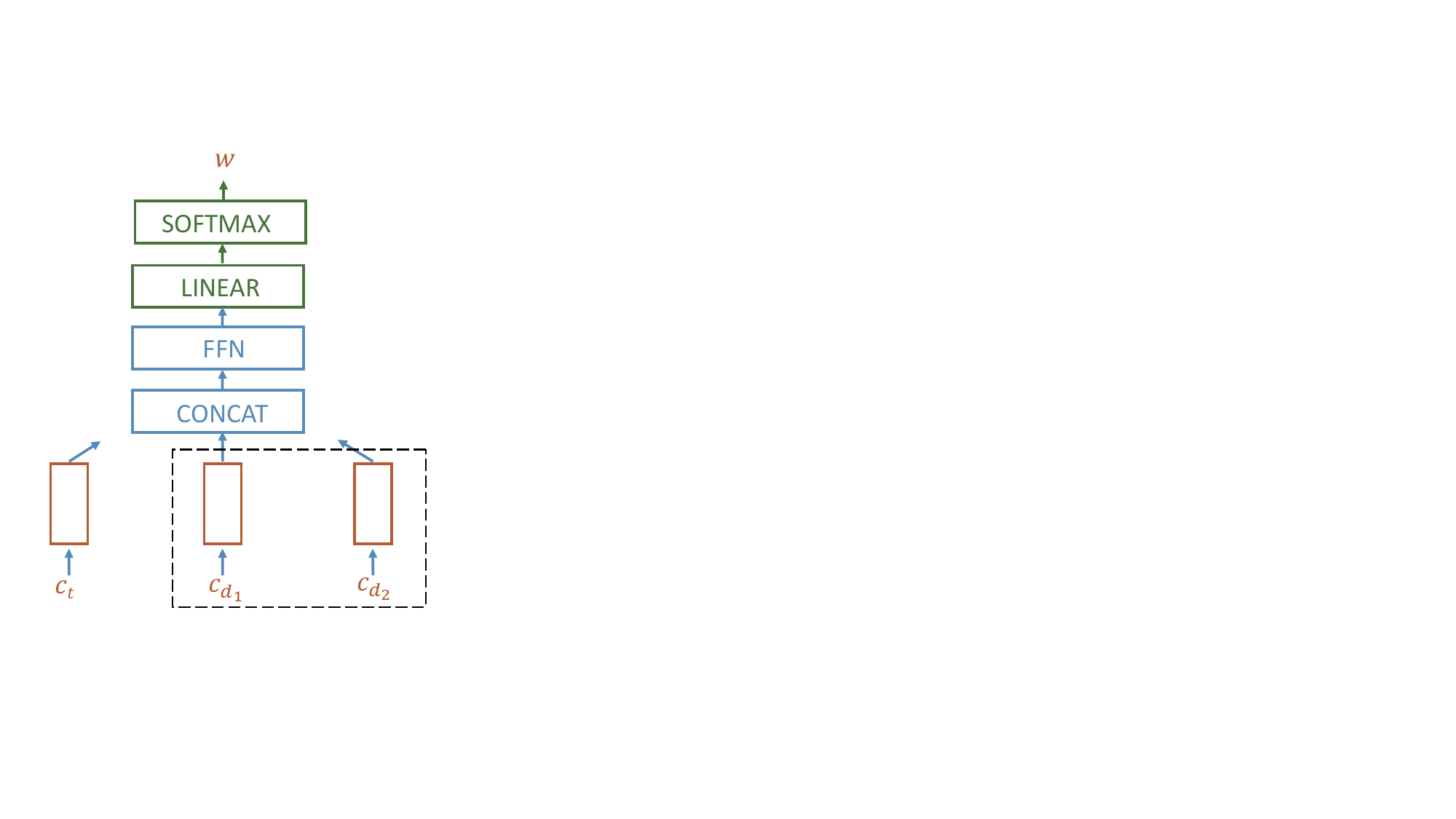}
    \caption{Agent Communication}
  \end{subfigure}
  \caption{Overview of agent architectures and the referential communication game. The speaking agent is presented with a target color ($c_t$) and, depending on the experimental condition, may also receive two distractor colors interpreted as context (dashed box). The listening agent always receives three colors and has to guess the index $i$ of the correct one ($c_t$), in this case 2. Both agents are first trained by SL on a dataset of human interactions.}
  \label{fig:agent_arch_comm}
\end{figure*}

The work by \citet{gualdoni-boleda-2024-objects} offers a promising bridge between these two traditions. They introduced a novel measure of informativeness for words and lexical systems that captures both context adaptation in language use and the structure of lexical systems. 
Meanwhile, \citet{lian2023communication} introduced the NeLLCom framework, which adopts a two-phase learning paradigm for simulating the emergence of syntactic universals \citep{lian-etal-2024-nellcom, lian2025simulating, zhang-etal-2024-endowing, zhang2024neural} without the need of complex, task-specific listener models typical of RSA.
In NeLLCom, agents are first trained to use a language with human-like structural properties via Supervised Learning (SL); then, those same agents are jointly optimized through interaction via Reinforcement Learning (RL). 
Building on these advances, we propose NeLLCom-Lex to simulate lexical change and measure lexical properties at both the language use and system level, using the metrics from \citet{gualdoni-boleda-2024-objects}. This enables a unified analysis of how communicative needs shape both in-context communication and lexicon structure.


\section{NeLLCom-Lex: A Neural-agent Framework for Lexical Systems}

Extending NeLLCom to simulate the evolution of lexical meaning via a referential game requires a number of non-trivial modifications.
In terms of meaning representations, NeLLCom-Lex replaces symbolic triplets with continuous perceptual inputs (i.e., CIELAB color vectors), providing a grounded representation space suitable for modeling lexical meaning in the color domain. Secondly, shifting from a meaning reconstruction task to a referential discrimination game 
requires architectural changes: Instead of processing utterance sequences with recurrent models, NeLLCom-Lex uses single-symbol utterances and incorporates a context encoder to model distractors. 
%
This section describes the resulting task and architecture.


\subsection{The task}

NeLLCom-Lex agents communicate about a simplified referential world using pre-defined lexicons acquired during SL.
Speaking agents are presented with a target color image \( c_t \) along with two distractor color images \( c_{d_1} \) and \( c_{d_2} \). The speakers' task is to generate a message (i.e., a color name \( w \)) to convey the meaning of the target color to the listener. 
Each word \( w \) is a discrete symbol drawn from a fixed vocabulary \( \mathcal{V} \), i.e., \( w \in \mathcal{V} \). Listening agents receive the message along with the same target color and distractors, presented in randomized order. 
A successful communication occurs if the listener correctly selects the target color. During SL, the message corresponds to a pre-defined color name; during RL, it is generated by the Speaker.
During RL, thus, agents play a referential discrimination game.

\subsection{Agent architectures}
Both speakers and listeners are composed of feedforward neural networks (FNNs), following the common architecture design in referential communication games \citep{chaabouni2021communicating, kharitonov2020entropy}.
Figure~\ref{fig:agent_arch_comm} shows the agent architectures and how they interact during communication.
For both kinds of agents, the target color chip (and, optionally, its distractors) is represented as CIELAB color vectors and mapped through feed-forward networks to obtain embeddings. These embeddings provide a grounded representation space for modeling lexical meaning in the color domain \cite{brainard2003color}. The color name \( w \) is mapped to its representation through an embedding layer. 

\paragraph{Speaker}
The speaker observes a referential context consisting of a target color \( c_t \) and two distractors \( c_{d_1} \), \( c_{d_2} \), with the target always in the first position. 
Each color is processed independently through its own feedforward block. The resulting embeddings are concatenated and passed through a fourth feedforward block, 
producing a joint representation used to predict a single-symbol message \( w \) via a final linear classifier.
To investigate how access to contextual information affects the speaker's lexical choices, we manipulate its input:
in the \textit{context-aware} setting, all three colors \( \{c_t, c_{d_1}, c_{d_2}\} \) are encoded normally; 
in the \textit{context-unaware} setting, distractor embeddings are zeroed out. 


\paragraph{Listener}

The listener receives the color name \( w \), along with the same referential context consisting of a target color \( c_t \) and two distractors \( c_{d_1} \), \( c_{d_2} \), presented in randomized order. 
Each of the three color candidates is independently projected into a hidden representation using a feedforward block. 
Simultaneously, the color name \( w \) is mapped to its representation. The listener then computes the similarity (dot product) between the embedding of \( w \) and each of the embedded color candidates.
These similarity scores are transformed into a log-probability distribution over the three candidate positions via a softmax layer. 
The listener is trained to select the position of the intended referent.

\subsection{Supervised language learning}

SL trains agents on a dataset \(\mathcal{D}\) of color triplets paired with target color names (see Section~\ref{sec:dataset} for details). 
The speaker learns to produce a name given a triplet, while the listener learns to identify the target color given a name and a shuffled triplet.

\paragraph{Speaker}

The speaker’s parameters \(\theta_S\) are trained by minimizing the cross-entropy loss:

\begin{equation}
\mathcal{L}_{\textit{S}}^{\text{sup}}(\theta_S) = -\sum_{(c, w) \in \mathcal{D}} \log p(w \mid c; \theta_S)
\end{equation}

\noindent where \(w\) is the ground-truth target color name, and \(c\) is the input color(s) provided to the speaker.

\paragraph{Listener}

The listener’s parameters \(\theta_L\) are optimized by minimizing the cross-entropy loss:

\begin{equation}
\mathcal{L}_{\textit{L}}^{\text{sup}}(\theta_L) = -\sum_{(w, \tilde{c}, i) \in \mathcal{D}} \log p(i \mid w, \tilde{c}; \theta_L)
\end{equation}

\noindent where \( \tilde{c} = \{\tilde{c}_1, \tilde{c}_2, \tilde{c}_3\} \) 
is the shuffled color triplet 
and \( i \in \{1, 2, 3\} \) the target color position in \( \tilde{c} \).


\subsection{Communication learning}

While speaker and listener models are trained independently during SL, they are updated jointly during RL to optimize communicative success. This phase allows us to explore how the learned lexical system adapts under pressure for communication efficiency. 
Following standard practice in emergent communication \citep{chaabouni2021communicating}, we simulate a referential game in which a speaker \(S\) conveys a target color \(c_t\) to a listener \(L\) by producing a symbol from a fixed vocabulary. In our setup, the speaker \(S\) conveys the target color using a word \(\hat{w}\) learned during SL.
The goal for both agents is to maximize a shared reward evaluated by the listener’s prediction. For this phase, we adopt the classical policy-based algorithm REINFORCE  \citep{williams1992simple}. Specifically, we optimize:

\begin{equation}
\mathcal{L}^{\text{comm}}_{\textit{S,L}} = -\sum \log p(\hat{w} \mid c; \theta_S) \cdot r^{L}
\end{equation}

\noindent where \(  r^{L} \) is the reward based on the listener’s prediction: $r^L = \log p(i \mid \hat{w}, \tilde{c}; \theta_L)$.


\section{Experimental Setup}

Our goal is to design agents that learn to use and adapt an existing lexical system in a way that mirrors human behavior.
This involves adapting word use to the communicative context (using more informative color names in hard contexts and less informative ones in easier contexts) at multiple scales: that is, situationally in synchronous communication, but also in the longer term by updating shared lexical conventions according to changes in the communicative needs of the environment.

We investigate the factors leading to more human-like behaving agents through two sets of experiments: different SL+RL training pipelines with varying access to context (\S\ref{sec:exp1}) and variations in the communicative environment of the agents during communication training (\S\ref{sec:exp2}), using the dataset and evaluation metrics described below. 

\subsection{Datasets}
\label{sec:dataset}
For SL, we use the English version of the \textbf{Colors} dataset processed by \citet{gualdoni-boleda-2024-objects} to analyze lexical properties, based on the original data collected by \citet{monroe-etal-2017-colors}.
In this data, human participants engage in a dyadic reference game, where
the speaker describes a target color chip from a grid containing two distractor chips. The goal is for the listener to correctly identify the target. 
Each target chip can appear in multiple contexts that vary in difficulty, defined by the visual similarity to distractors:
\textsc{\textbf{far}} (both distractors clearly distinct from the target), 
\textsc{\textbf{split}} (one distinct and one similar distractor), 
\textsc{\textbf{close}} (both distractors similar to the target).
The dataset contains all the successful communication rounds solved with a single word, and comprises 9,309 far, 3,886 split, and 2,239 close instances (15,434 in total). 

To ground agents' language use, SL is conducted using explicit lexical labels from the \textbf{Colors} dataset. Of these, 3K~instances are held out for SL testing (\textsc{test}$_{hum}$), while the remaining 12.4K 
are used for SL training (\textsc{train}$_{hum}$).

For RL training and evaluation where human labels are not needed, we generate a large number of color triplets for far/split/close context types, following the sampling procedure described by \citet{monroe-etal-2017-colors} (see Appendix \ref{app:appendix_data_model}, for data generation and model training details). The generated test set distributions differ by experiment and are described in \S\ref{sec:exp1} and \S\ref{sec:exp2} respectively.
This allows us to scale up training without requiring additional labeled data, since RL only relies on communicative feedback.

\subsection{Evaluation} 
Agent productions are always 
evaluated on a held-out test set of contexts
(i.e., combinations of target and distractor colors)
that are not seen during any training phase.
To ensure robust results, all experiments are repeated with 10 different 
random seeds and the averaged evaluation metrics are reported.  

\paragraph{Accuracy} 
To assess the performance of the SL phase, we measure \textbf{speaking} and \textbf{listening accuracy} ($Acc_{spk}$ and $Acc_{lst}$ respectively) against the human labels provided in the \textbf{Colors} dataset. These measures capture the extent to which agents have learned to master a given lexical system $L$.
\textbf{Communication accuracy} or \textbf{success} ($Acc_{comm}$) is instead defined as the proportion of trials where the listener
successfully identifies the correct color label based on the input received from the speaker. 

\paragraph{Pragmatic adaptation} Recall that humans adapt their naming choices to the difficulty of the referential task. For instance, in the top row, right column of Table~\ref{tab:color_naming}, we see that the same chip was called ``green'' when appearing next to a gray and a blue chip, but ``sage'' when it appeared next to two other green chips of different shades, as ``green'' would be ambiguous in this context.
To test whether agents also learn this kind of pragmatic adaptation, we follow \citet{gualdoni-boleda-2024-objects} and test whether more \textbf{informative} words are used in more referentially difficult contexts, the two quantities being defined as follows.
\textbf{Word informativeness} ($I_w$) is measured by grounding color labels in the set of color chips they denote, placing these referents within a convex visual space.
More specifically,
given word $w$, the spread of its visual features ($S_w$) is defined as the average pairwise distance between the objects \emph{o} that are referred to by \emph{w}: 
\begin{equation}
    S_w=\frac{1}{N}\sum_i\sum_{j\neq i}d(o_i,o_j)
\end{equation}
where \emph{N} denotes the number of unique object pairs, and $d(o_i,o_j)$ refers to the Euclidean distance between objects $o_i$ and $o_j$ in the CIELAB space, when both objects are denoted by the same word \emph{w}. 
The inverse of the spread gives us word informativeness: $I_w=1/S_w$.
Thus, words that apply to smaller subsets of color chips are considered more informative about the denoted shade and will have higher $I_w$ values. 
\textbf{Context ease} ($E_{ctx}$) is defined as the distance between the target color chip and the hardest distractor in the context (i.e. the distractor being closest in CIELAB color space). 

To examine the relationship between word informativeness and context ease, we fit a linear mixed-effects model predicting $I_w$ for the word generated by the speaker, using $E_{ctx}$ as a fixed factor.
Agent initialization seeds and target chips are treated as random effects.\footnote{We consider only the target chips that appear at least twice in the dataset, following \citet{gualdoni-boleda-2024-objects}.}
\label{sec:ctx_lme}

\paragraph{Properties of the lexicon}
Finally, we define three system-level metrics to examine the properties of the agents' lexicons. \textbf{System-level informativeness} \cite{gualdoni-boleda-2024-objects} is defined as the average informativeness $I_w$ of the words used to solve \emph{N} interactions.
\begin{equation}
    I_L=\frac{1}{N}\sum_{i=1}^NI_w^i
\end{equation}
It captures the average informativeness afforded by a given lexicon. Human-like lexicons have mid $I_L$ values---enough to make useful distinctions, without being unnecessarily informative \cite{gualdoni-boleda-2024-objects}.
Following \citet{regier2015word}, we use a different measure of lexical complexity, namely the number of different color words, or word types, used by a speaking agent evaluated on a given set of color chips ($|W|$). We call this \textbf{lexical diversity}.
The third measure, \textbf{semantic drift}, quantifies the divergence of the agents' lexicons from the human lexicon by computing the distance between their word \textbf{prototypes} \cite{gualdoni_whats_2023}. First, the prototype of each color name is obtained by averaging the CIELAB coordinates of all color chips labeled with that name.
Then, for every color used by the agents, we compute the Euclidean distance
between its prototype and the corresponding human prototype.
We define the semantic drift ($D_L$) of a lexical system $L$ as the average of these distances.

\section{Experiments and Results}

We carry out two sets of experiments. The first set (\S\ref{sec:exp1}) aims at testing the conditions under which agents develop pragmatic naming behavior and lexical properties that are similar to those of humans. In particular, we check how and when access to context is necessary for human-like lexicons to emerge. 
The second set (\S\ref{sec:exp2}) tests the effect of different communicative needs on the resulting behavior and lexicons; in particular, it exposes agents to more and less granular color distinctions.

\subsection{How does context access during training affect agent behavior?}
\label{sec:exp1}



We set up different SL and RL training pipelines to assess the extent to which context-sensitive training leads agents to acquire more or less human-like color naming behavior, based on the metrics defined in the previous section.
We also study the effect of introducing context in different phases of the pipeline. For example, \textbf{SL$+$} refers to only SL with access to context, while \textbf{SL$-$RL$+$} means SL without context followed by RL with context.\looseness=-1

In these experiments, RL training and evaluation are performed on generated data matching the context ease distribution of the human \textbf{Colors} dataset which is used for SL training. Specifically, we use 12.4K 
color triplets for training (\textsc{train}$_{gen,distH}$) and 15.4K 
for testing (\textsc{test}$_{gen,distH}$). 
We report agent productions on the overall (O) test set, as well as separately on far (F) and close (C) contexts.\footnote{
Split contexts are part of this test set, but we do not report their performance separately as it highly correlates with that of close contexts,
as reported in \citet{gualdoni-boleda-2024-objects}.
}

\begin{figure}[t]
  \centering
  \begin{subfigure}[b]{0.325\linewidth}
    \centering
    \includegraphics[width=\linewidth]{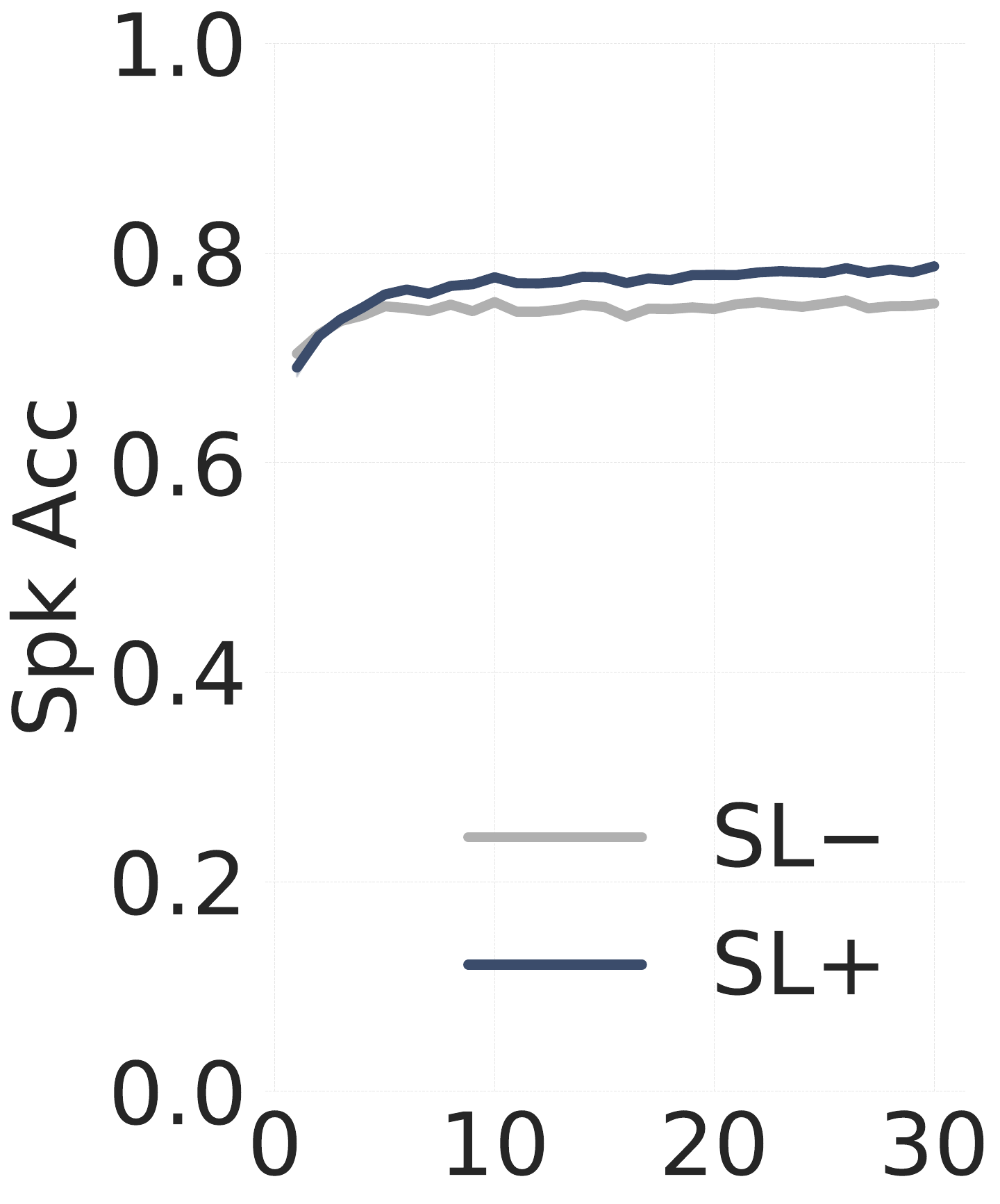}
    \label{fig:acc_spk_human_dist}
  \end{subfigure}
  \begin{subfigure}[b]{0.325\linewidth}
    \centering
    \includegraphics[width=0.97\linewidth]{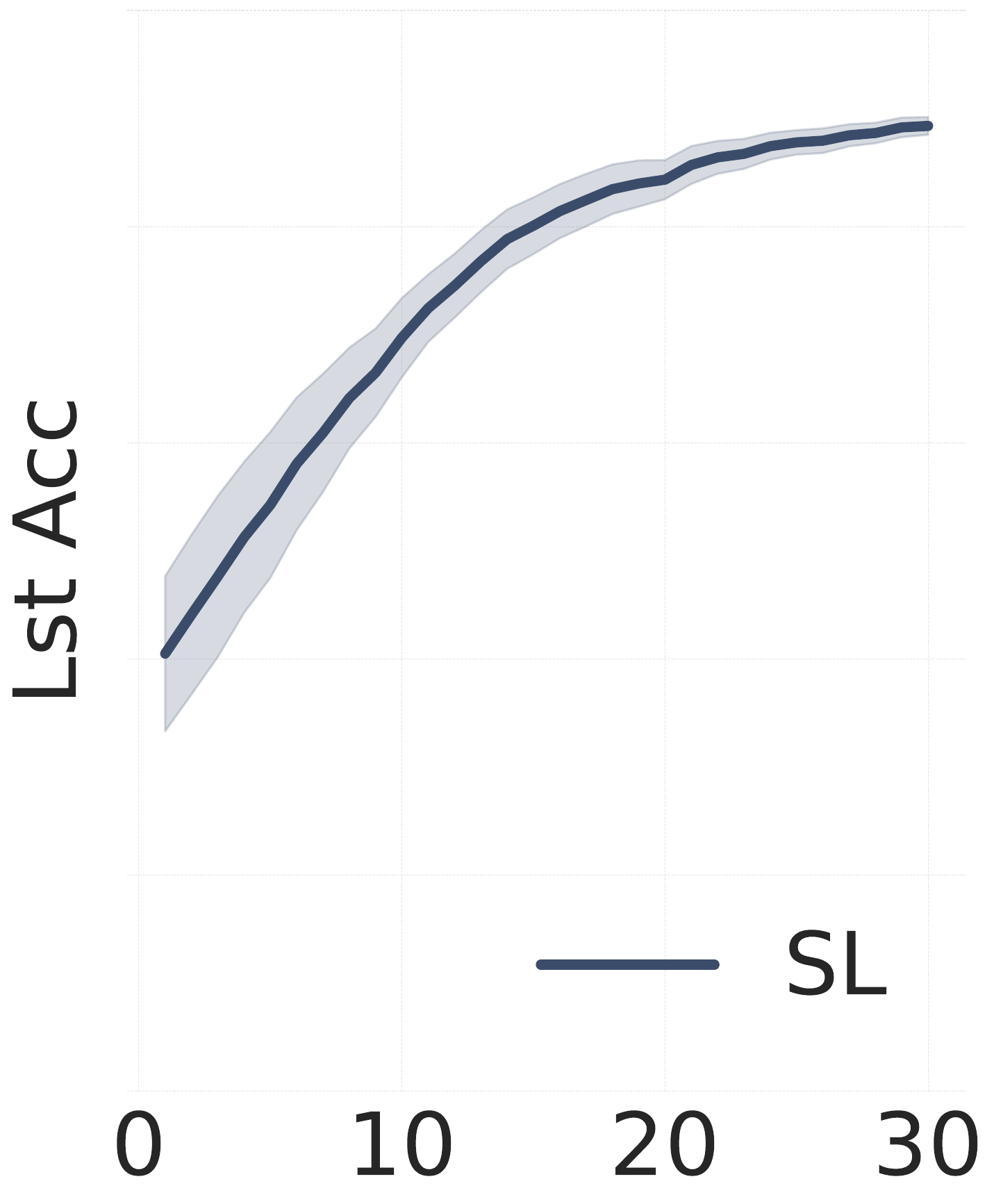}
    \label{fig:acc_lst_human_dist}
  \end{subfigure}
  \begin{subfigure}[b]{0.325\linewidth}
    \centering
    \includegraphics[width=0.97\linewidth]{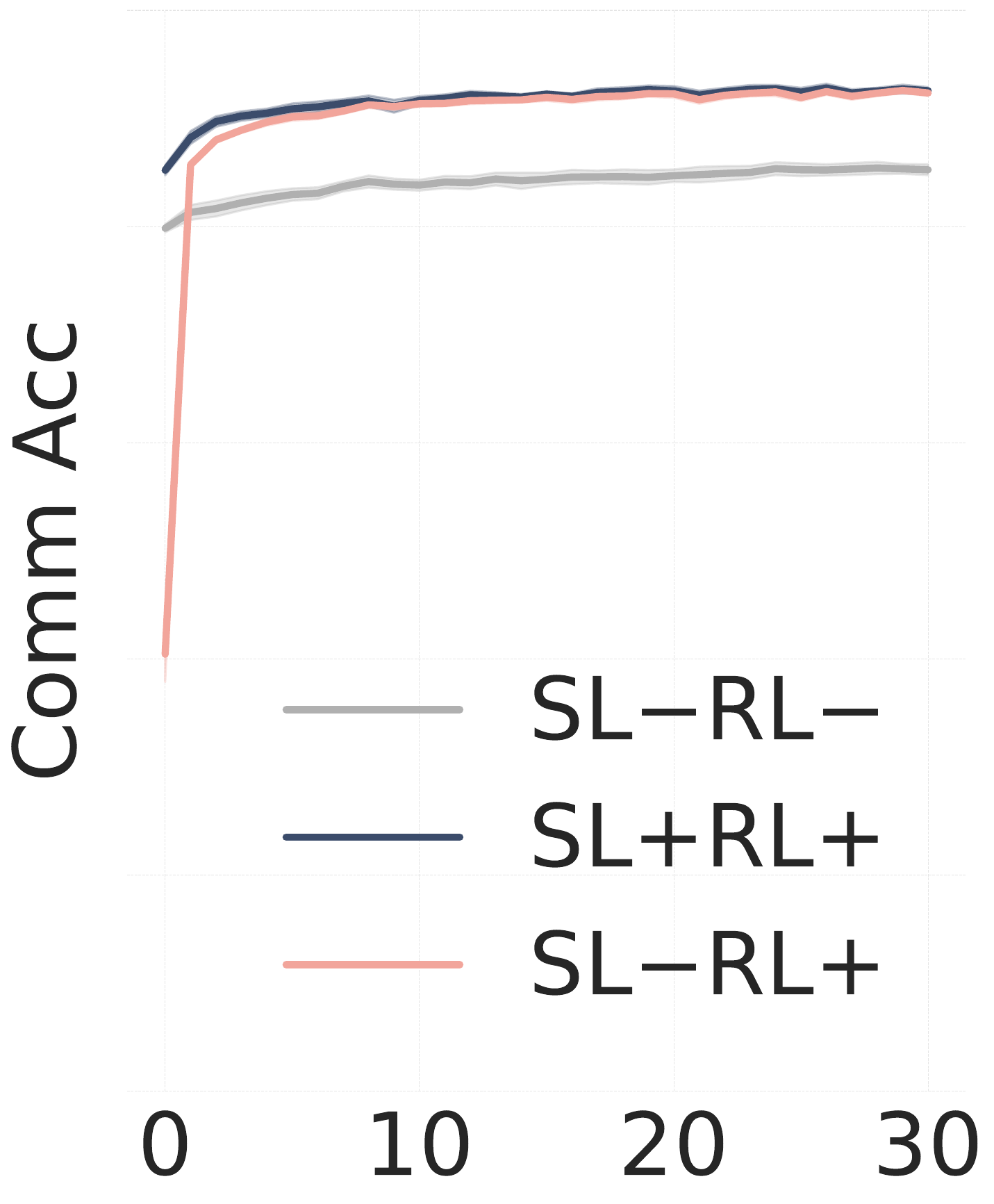}
    \label{fig:acc_comm_human_dist}
  \end{subfigure}
  \caption{Mean speaking and listening accuracy on \textsc{test}$_{hum}$, and mean communication accuracy on \textsc{test}$_{gen,distH}$ as a function of training epochs across different training pipelines. In the right-most plot, epoch~0 denotes the state immediately after SL and before RL.
  Shaded regions indicate 95\% confidence intervals. All values are averaged over 10 random seeds.}
  \label{fig:acc_combined}
\end{figure}

\paragraph{Accuracy}
Figure \ref{fig:acc_combined} shows agents' speaking and listening accuracy on the human-labeled \textsc{test}$_{hum}$, and communication accuracy on the generated \textsc{test}$_{gen,distH}$. 
Speaking agents achieve slightly higher accuracy when trained with context (79\% vs. 75\% without context), suggesting the benefit of contextual information to approximate human naming data.\footnote{Speaking agents cannot reach 100\% accuracy since a single target color chip can be mapped to multiple color names in the dataset because of the naming variation across participants.} Listener agents reach 90\% accuracy by epoch 30, with no differences across conditions, as context only affects the speaker.
As for \textbf{communication accuracy}, agents trained with context during both SL and RL start below 90\% when evaluated on the generated dataset right after SL training (epoch 0), indicating limited generalization to the novel dataset despite pragmatic behavior learned during SL. 
${\rm Acc}_{comm}$ rapidly improves after just one epoch of RL, 
suggesting that access to context during communication-based training is needed for pragmatic language use in novel situations. 
Agents trained without context in SL but with context in RL initially struggle to communicate 
but quickly reach comparable accuracy to fully context-trained agents. 
This suggests that agents can learn to use context effectively regardless of whether they were trained with it in the SL phase. 
By contrast, and as expected, agents trained without context in both SL and RL achieve the lowest accuracy, highlighting the necessity of contextual exposure for robust communication.

Next, we study how the different pipelines affect the agents' 1)~pragmatic naming behavior, 2)~lexicons (informativeness, lexical diversity, and semantic drift).
The results are summarized in Table~\ref{tab:parta}.\looseness=-1

\begin{table}[h!]
\centering \small
\begin{tabular}{lcccc}
\toprule
\textbf{Training} & \boldmath{$\beta(E_{ctx})$} & \boldmath{$I_L$} & \boldmath{$|W|$} & \boldmath{$D_L$} \\
\midrule
SL$-$              & 0.000          & 3.59 & 8.50 & 14.7 \\
SL$+$              & \boldmath$-0.003$ & 3.23 & 12.7 & 20.3 \\
SL$-$RL$-$         & $-0.000$        & 5.28 & 21.1 & 46.2 \\
SL$-$RL$+$         & \boldmath$-0.002$ & 2.53 & 32.6 & 48.4 \\
SL$+$RL$+$         & \boldmath$-0.002$ & 2.60 & 34.5 & 47.3 \\
\midrule
Human              & \boldmath$-0.008$                & 2.78 & 49.0 & -- \\
\bottomrule
\end{tabular}
\caption{Agents' production properties on \textsc{test}$_{gen,distH}$. 
$\beta(E_{ctx})$: estimated effect of context ease on word informativeness with significant values in boldface ($p$$<$0.001; standard errors for all estimates are below 0.001). ; 
$I_L$: system-level informativeness; $|W|$: lexical diversity; $D_L$: semantic drift.
Human values are reported from \citet{gualdoni-boleda-2024-objects} and refer to the full 15.4K \textbf{Colors} dataset.\looseness=-1
}
\label{tab:parta}
\end{table}

\paragraph{Pragmatic adaptation}

\begin{figure}[t]
  \centering
    \includegraphics[width=\linewidth]{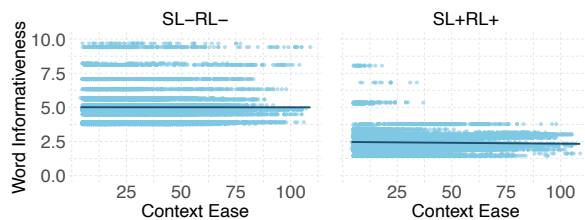}
  \caption{Word informativeness as a function of context ease on \textsc{test}$_{gen,distH}$ for 2 representative seeds. 
  }
\label{fig:informativeness_scatter_parta}
\end{figure}

Table~\ref{tab:parta} and Figure~\ref{fig:informativeness_scatter_parta} show that neural agents indeed exhibit pragmatic adaptation to context if they are exposed to the referential context at training time.
$I_w$ decreases slightly but significantly as context becomes easier for agents exposed to context, either during SL, 
RL 
or both. 
Notably, there is no significant difference between the three context-aware pipelines.%
\footnote{See Appendix \ref{app:comparison_parta} for all pairwise statistical testing results for the current and subsequent modeling. Additional scatterplot examples are provided in Appendix~\ref{app:informativess_scatter_parta}.}
This indicates that access to context leads agents to develop human-like pragmatic behavior regardless of whether it is introduced during SL or RL.
The latter does matter for other aspects, as we will see below.

\paragraph{Properties of the lexicon} 
Agents without access to context not only fail to exhibit pragmatic behavior, but also develop inefficient lexicons: 
On the one hand, they are overly informative ($I_L$=3.59 and 5.28 for SL$-$ and SL$-$RL$-$, respectively, compared to 2.78 for humans); on the other hand, despite this high informativeness, they are not as successful at communicating as the others (see Figure~\ref{fig:acc_combined}).
This pattern is also visible in Figure~\ref{fig:informativeness_scatter_parta} (for further analyses of informativeness distribution, see Appendix \ref{app:entropy_informativeness}). 
These agents also develop very poor lexicons (8.5 and 21.1 average word types, compared to 49 for humans).
Crucially, even if agents develop pragmatic behavior already after supervised training if they have access to context (SL+ condition), that doesn't mean that their lexicons are human-like: they are still over-informative ($I_L$ 3.23) and, especially, they are very poor, with only 12.7 word types on average.
Instead, allowing both access to context and interaction while learning (SL+RL+) yields the closest lexicon to the human one, with $I_L$ 2.6 and vocabulary size 34.5.


These patterns suggest that access to context and communication not only drive pragmatic behavior but also encourage the development of a richer, more adaptive lexicon (see Appendix \ref{app:lexical_diversity_parta} for analysis across test subsets, i.e., far and close contexts).

The last column of Table~\ref{tab:parta} reports the average semantic drift, measured as the divergence between model and human prototypes.\footnote{See also Appendix~\ref{app:word_embeddings} for a visualization of the speaker's word embeddings before and after RL.}
As could be expected, drift leaps after RL in all conditions. Thus, a trade-off appears between keeping words grounded in human language and maintaining a lexical system with overall human-like properties.

\paragraph{Interim conclusion}
From this set of experiments, we conclude that the overall NeLLCom pipeline, where agents first acquire a given linguistic system through supervised learning and subsequently adapt it via reinforcement learning, shows promise to model the dynamics between language use and the properties of lexical systems. The setting that yields the most human-like behavior is SL+RL+ (i.e., where agents have access to context both at SL and at RL time). In this setting, agents develop lexical systems that are at a middle level of informativeness and remarkably varied (both properties observed in humans), and they exhibit pragmatic adaptation to context in synchronic production (again as observed in humans).

\subsection{Changing communication needs}
\label{sec:exp2}

We hypothesize that agents will adapt their behavior and lexicons to varying referential needs.
To test this, we vary the distribution of contexts during RL training, exposing agents to circumstances where they need to make more or less granular color distinctions (e.g., do they need to identify a given turquoise chip among red and yellow chips, or among aqua and cyan?).
In these experiments, we adopt the context-aware pipeline (SL$+$RL$+$) and train agents in \textbf{AllFar}, \textbf{HalfHalf}, and \textbf{AllClose} context conditions, with 15.4K training instances for each condition.
%
We also generate a balanced 15.4K-instances test set (\textsc{test$_{gen,dist50}$}) comprised of 50\% far and 50\% close contexts, where each target color appears once in each context type. 
Importantly, the test set is kept fixed across the three different RL training distributions.
However, we report test performance overall (O) as well as separately on far (F) and close (C) contexts.

In terms of pragmatic adaptation, we expect AllClose-trained agents to develop a preference towards more specific naming overall, 
HalfHalf-trained agents to maintain or even amplify distinctions in naming between close and far test subsets,
and AllFar-trained agents to reduce the pragmatic distinctions.
As for lexical properties, we expect lexicons of HalfHalf agents to be closest to human-like lexicons among the three conditions, with a varied lexicon at a mid level of informativeness.
Lastly, we expect lexicons of AllFar agents to be the least flexible according to these metrics (and insufficient to properly solve close contexts).

\begin{figure}[h]
  \centering
    \includegraphics[width=\linewidth]{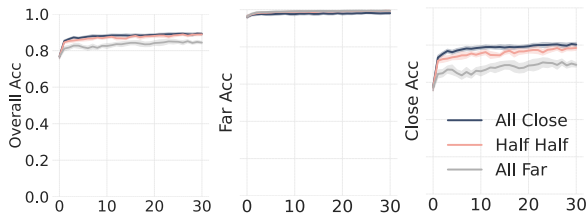}
  \caption{Communication accuracy on \textsc{test$_{gen,dist50}$} for varying RL training distributions, as a function of RL training epoch. Each plot shows accuracy on a different test subset: Overall (O), Far (F), and Close (C).
  } 
  \label{fig:accuracy_plot_partb}
\end{figure}

\paragraph{Accuracy} The overall accuracy results (Figure~\ref{fig:accuracy_plot_partb}) match our expectations: all agents are almost perfect at far contexts, but accuracy in close contexts is much lower (\textasciitilde70-80\%). Not only AllFar, but also HalfHalf agents struggle more with close contexts than AllClose.
This suggests that communication training is especially useful to negotiate specific terms to be used in difficult contexts, whereas less informative terms are already well mastered by agents at the end of SL.


\paragraph{Pragmatic adaptation}
As shown in Table~\ref{tab:informativeness_partb}, agents exhibit pragmatic adaptation in all three conditions. 
%
%
Moreover, a significant interaction between context ease and RL training condition 
confirms that this effect varies across conditions. 
Specifically, agents trained exclusively on difficult contexts 
show slightly stronger context sensitivity. In contrast, agents trained on easy contexts show 
less context sensitivity, 
reflecting their reduced need to modulate informativeness by context ease. 
Figure 5 
illustrates how this may affect the shift in informativeness of individual colors. The visual space covered by the label \emph{pink} varies across RL conditions: in the AllFar condition, the label 
\emph{pink} spans a broader region in the CIELAB space, corresponding to a lower informativeness. In AllClose, it denotes a much narrower area, and higher informativeness (see Appendix \ref{app:3d_cielab} for more examples).
The AllClose agents exhibit the greatest context sensitivity, followed by the HalfHalf agents and, finally, the AllFar agents. This suggests that dealing with more specific contexts may be more important than being exposed to qualitatively different contexts, though further tests are needed to probe this hypothesis.

\begin{table}[h!]
\centering \small
\begin{tabular}{p{3.5cm}c}
\toprule
\textbf{Conditions} & \textbf{\boldmath$\beta(E_{ctx})$} \\
\midrule
Before RL     & \textbf{-0.007} \\ 
\midrule
AllClose     & \textbf{-0.005} \\
HalfHalf     & \textbf{-0.004} \\
AllFar       & \textbf{-0.003} \\
\bottomrule
\end{tabular}
\caption{Effect of context ease on word informativeness across RL training distributions, all measured on \textsc{test$_{gen,dist50}$}. All effects are significant ($p$$<$.001).
}
\label{tab:informativeness_partb}
\end{table}

\begin{figure}[t]
  \centering
  \begin{subfigure}[b]{0.32\linewidth}
    \centering
    \includegraphics[width=\linewidth]{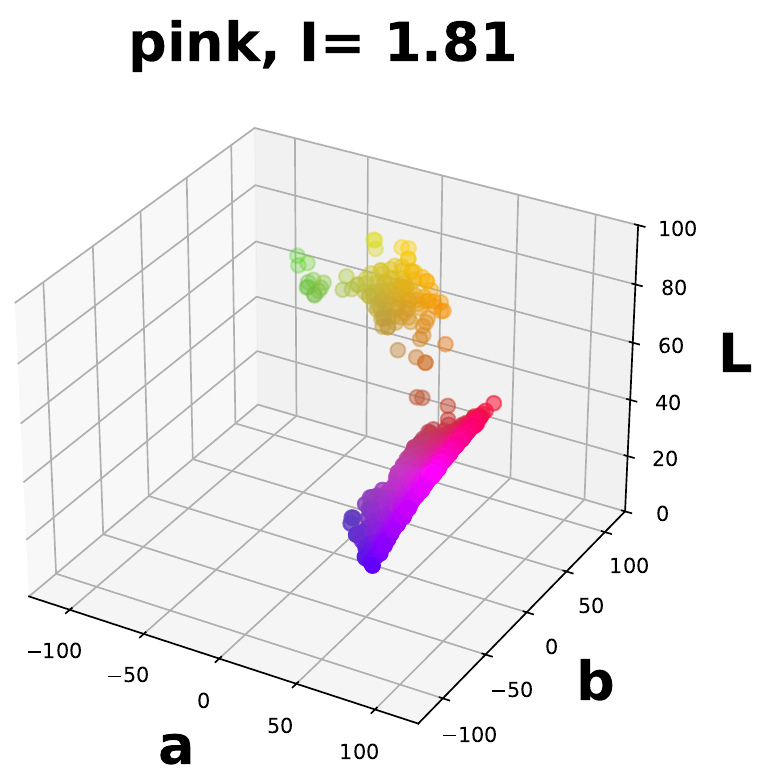}
    \caption{AllFar}
    \label{fig:far_example}
  \end{subfigure}
  \begin{subfigure}[b]{0.32\linewidth}
    \centering
    \includegraphics[width=\linewidth]{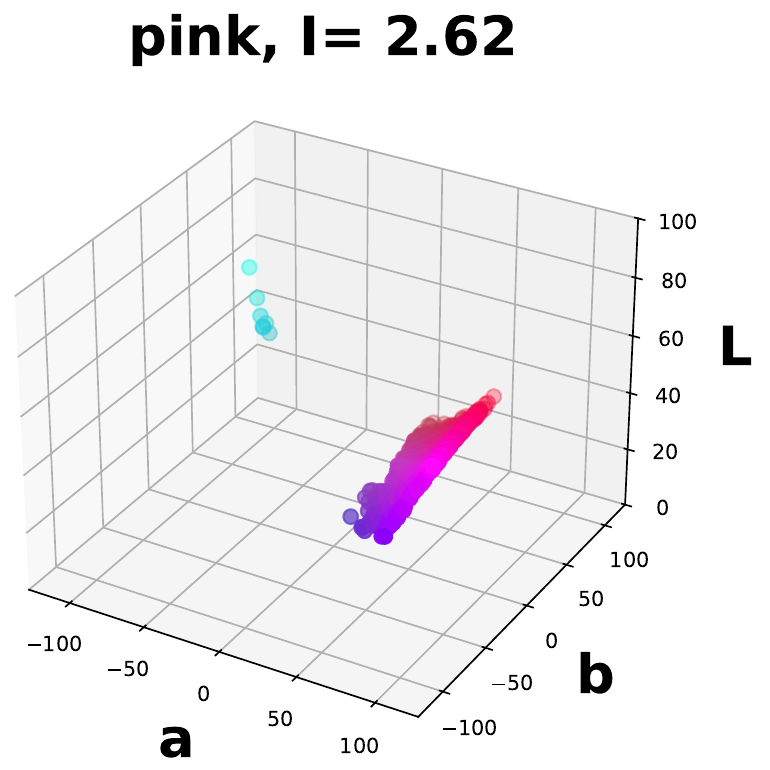}
    \caption{HalfHalf}
    \label{fig:half_example}
  \end{subfigure}
  \begin{subfigure}[b]{0.32\linewidth}
    \centering
    \includegraphics[width=\linewidth]{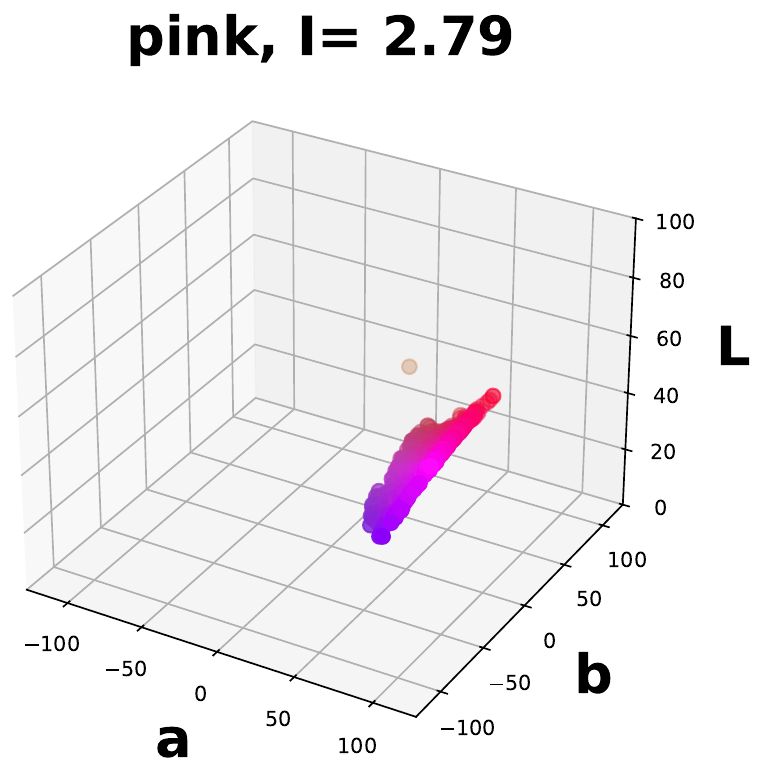}
    \caption{AllClose}
    \label{fig:close_example}
  \end{subfigure}
\caption{Denotation in the CIELAB color space for \textit{pink} in the three RL training conditions.}
\label{context_example}
\end{figure}

\begin{figure}[h]
  \centering
  \includegraphics[width=\linewidth]{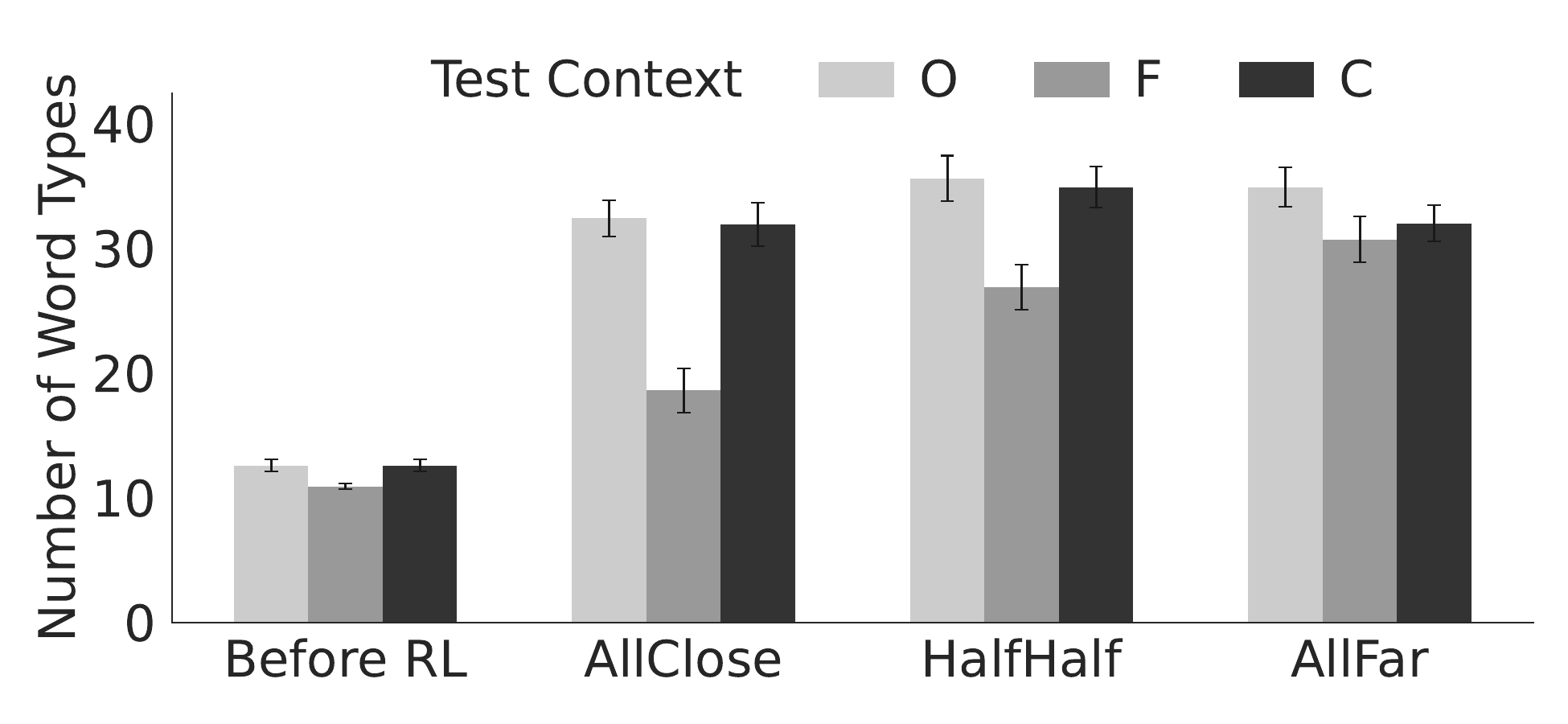} 
  \caption{Number of distinct word types used before RL and after RL training on varying context conditions (O: overall; F: far; C: close). }
  \label{fig:word_type_usage_partb}
\end{figure}

\paragraph{Properties of the lexicon}
The results above suggest that the more human-like lexicons 
are those of AllClose, followed by HalfHalf and AllFar. We see further evidence for this in Figure~\ref{fig:word_type_usage_partb}, where AllClose agents need a much smaller vocabulary while reaching the highest accuracy (Figure \ref{fig:accuracy_plot_partb}).
This suggests that their lexicons are more efficient.
The clearest contrast is AllFar, where the gap between diversity in close vs.\ far contexts narrows and almost disappears.
This suggests that increased exposure to far contexts leads agents to generalize their lexicon uniformly across context types, rendering it inefficient.
However, counter to expectation, we find that the overall lexical diversity is very similar across conditions, 
as are system-level informativeness scores (see Appendix~\ref{app:lexical_diversity_parta} and \ref{app:system_informativeness_partb}). 
The informativeness and lexical diversity values are similar to those of \S\ref{sec:exp1}, so reasonably close to those of humans. We do not find evidence for a comparatively coarse-grained lexicon for AllFar agents.

Summing up, we find that training in a variety of contexts allows agents to develop more flexible pragmatic behavior and adaptive lexicons. Still, further work is needed to fully understand the relationship between the referential context and the properties of the lexicon.


\paragraph{Pragmatic naming examples}
Grounding agent language in a human lexical system enables direct comparison with human behavior. 
Table~\ref{tab:color_naming} provides some representative examples evaluated on \textsc{test}$_{hum}$, showing that, in some cases, SL-trained agents already use broader terms (e.g. \textit{green}) in far contexts, but prefer more informative labels (\textit{olive}) in contexts where finer distinctions are needed, similarly to the human reference labels (\textit{green}-\textit{sage}). 
In other cases, however, the same term (\textit{purple}) or a comparably broad term (\textit{red}-\textit{brown}) is used in both context types.
After RL, the pragmatic naming behavior of the agents becomes more consistent (e.g. \textit{red} vs. \textit{teal} or \textit{concrete}).
At the same time, the use of unexpected terms, like \textit{gray} in the first color example, reveals that semantic drift has occurred.

\begin{table}[h!]
\centering \small
\resizebox{\linewidth}{!}{%
\begin{tabular}{cccc}
\toprule
& \textbf{Agent 1} & \textbf{Agent 2} & \textbf{Human} \\
\midrule
\includegraphics[width=1.5cm, height=0.5cm]{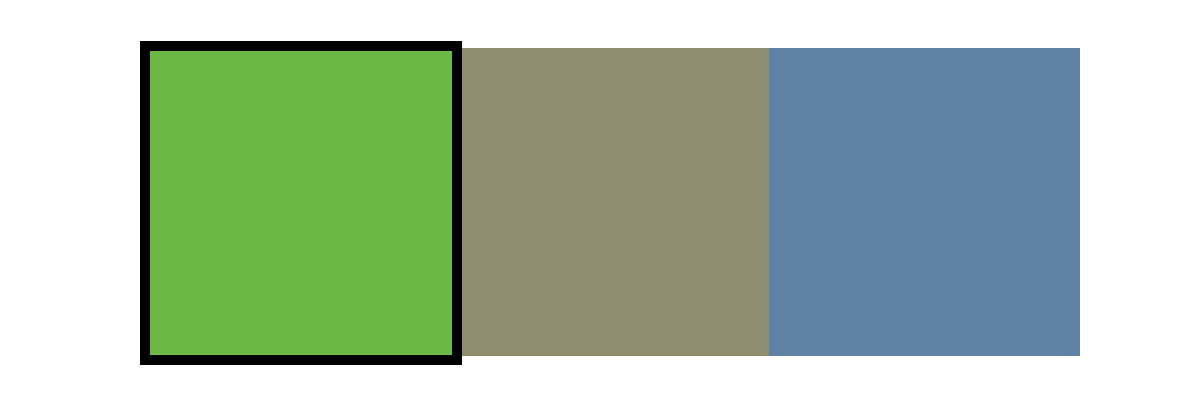} & green $\rightarrow$ green & green $\rightarrow$ green & green \\
\includegraphics[width=1.5cm, height=0.5cm]{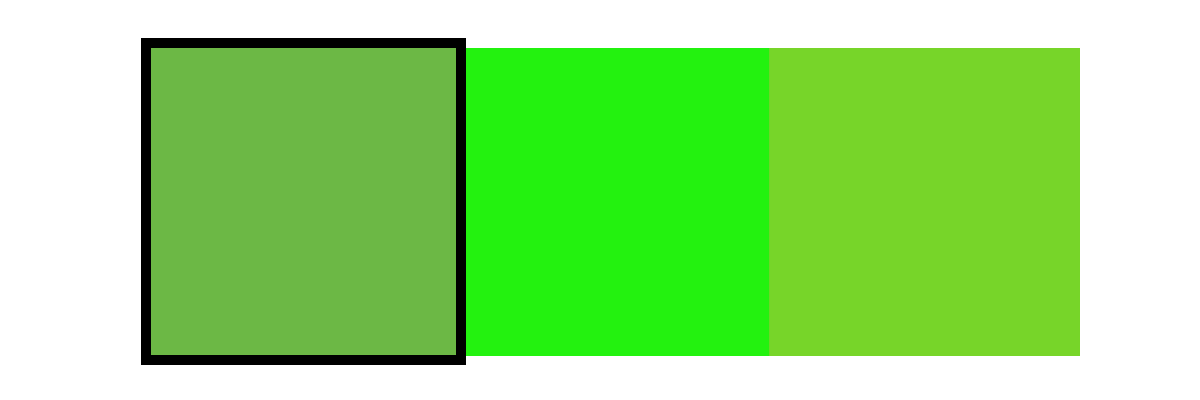} & olive $\rightarrow$ purple & olive $\rightarrow$ gray & sage \\
\midrule
\includegraphics[width=1.5cm, height=0.5cm]{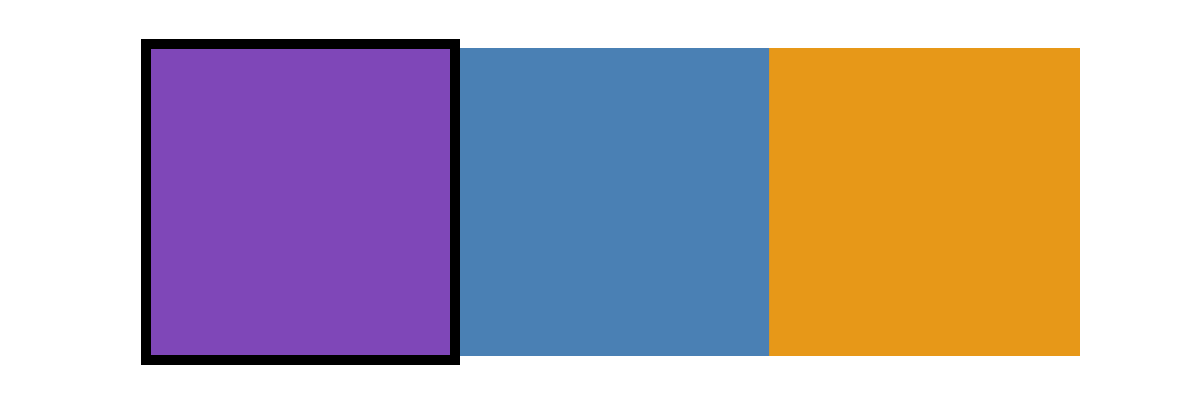} & purple $\rightarrow$ purple & purple $\rightarrow$ purple & grapes \\ 
\includegraphics[width=1.5cm, height=0.5cm]{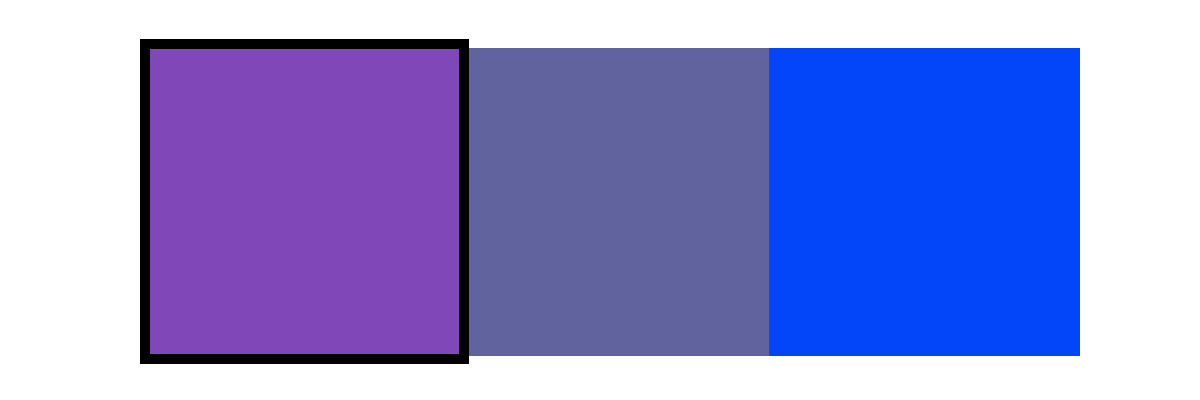} & purple $\rightarrow$ pink & purple $\rightarrow$ red & purple \\
\midrule
\includegraphics[width=1.5cm, height=0.5cm]{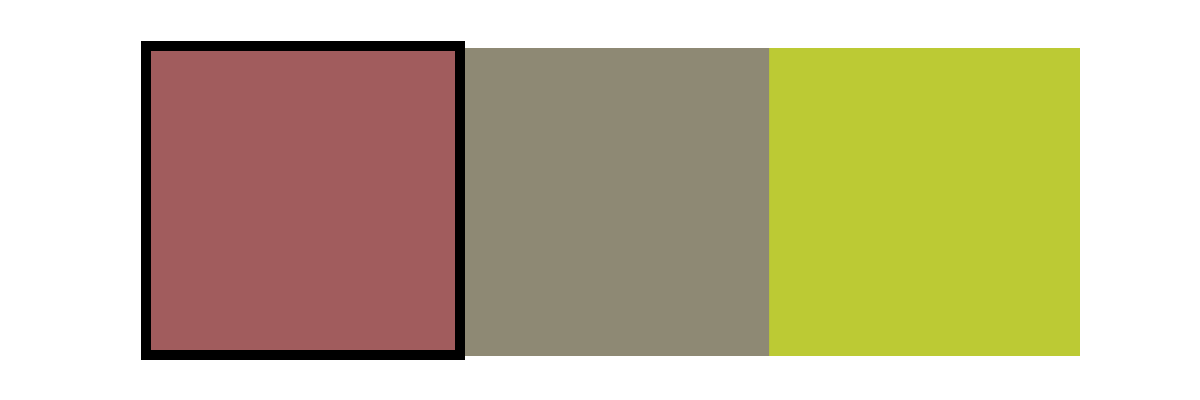} & red $\rightarrow$ red & red $\rightarrow$ red & red \\
\includegraphics[width=1.5cm, height=0.5cm]{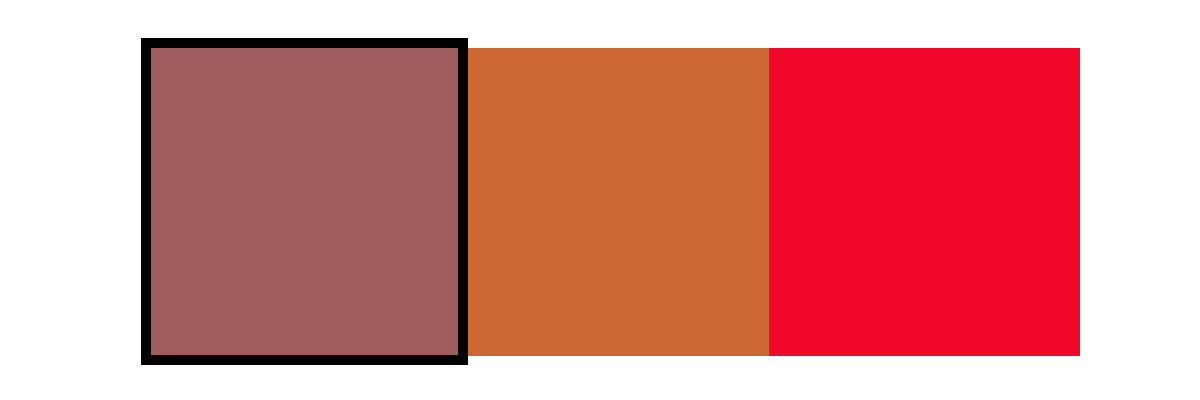} & brown $\rightarrow$ teal & brown $\rightarrow$ concrete & mauve \\
\bottomrule
\end{tabular}%
}
\caption{Color naming examples from two speaking agents, along with the human reference label. In each of the three examples, the same target color (in the black frame) is presented in far (top) vs. close (bottom) contexts. Arrows indicate naming changes: before$\rightarrow$after RL training in the AllClose condition.} 
\label{tab:color_naming}
\end{table}

\section {Discussion and Conclusion}
We introduced NeLLCom-Lex, a neural-agent framework to study the interplay between language use and the evolution of lexical systems.
Using a well-established color naming task for which human naming data is available, we examined which factors lead to more 
human-like pragmatic naming behavior in agents initially trained to `speak' an existing lexical system (in our case: English).
We found that agents learn to adapt their word use to context (i.e., using more informative terms in harder contexts and less informative ones in easier contexts) regardless of whether context is provided to them 
during supervised learning,
communication learning, or both. 
Furthermore, we found that optimizing communication accuracy through a referential game led the agents to develop a richer, more adaptive lexicon. 
In additional experiments, we simulated changing communicative needs in the agents' environment by manipulating the distribution of different context difficulties during RL training.
Results show that agents trained mostly 
in close contexts developed stronger pragmatic adaptations than those trained in far contexts, supporting the idea that pragmatic efficiency is driven by communicative needs.

Overall, our experiments show that neural agents can reproduce human-like patterns in color naming behavior, supporting their use as models for studying how language evolves under communicative pressures. 
In future work, we plan to use NeLLCom-Lex to simulate specific instances of lexical semantic change, such as meaning narrowing and broadening, that have been traditionally assumed to be driven by changes in speakers' communicative needs, but whose mechanisms remain to be explored \cite{Bréal1897,Geeraerts2020semantic-change}.



\section*{Limitations}

While our study sheds light on the conditions under which pragmatic behavior emerges in artificial neural agents, it also leaves open several questions. For example, our agents interact within a relatively constrained referential space (color triplets), which may limit the generalization of the findings to more complex semantic domains.

During SL, the training data is not uniformly distributed across the three context types (far, split, close). This imbalance may bias the agents toward overfitting to the more frequent context types, while disadvantaging learning in the close condition, where referential contrast is most challenging. As a result, the observed trend may underestimate the agents' capacity for pragmatic adaptation.

It is also worth noticing that RL training is affected by the entropy regularization parameter, which regulates the degree of exploration and encourages a more dispersed output distribution from the agent. Previous work has shown that overall message entropy after communication training tends to remain near the lower bound, even under strong regularization \citep{kharitonov2020entropy}. Future work could explore how varying the entropy regularization parameter influences agents' pragmatic naming behavior and contributes to the shaping of their emergent lexicon.

Finally, the current study models only dyadic interactions and lexical evolution across a single generation. 
For the latter, agents would need to be iteratively trained with the previous generation of agents' productions \citep{carlsson2024cultural}.
Future work could also study group-level dynamics, such as role alternation 
and group interaction patterns, following recent developments in agent communication \citep{michel2023revisiting, lian-etal-2024-nellcom}. 

\section*{Acknowledgments}
Arianna Bisazza was partially funded by the Dutch Research Council (NWO) within the Talent Programme (VI.Vidi.221C.009) and the InDeep project (NWA.1292.19.399). Gemma Boleda was partially funded by grant PID2020-112602GB-I00/MICIN/AEI/10.13039/501100011033, of the Ministerio de Ciencia e Innovación and the Agencia Estatal de Investigación (Spain).

\bibliography{custom}

\begin{thebibliography}{50}
\providecommand{\natexlab}[1]{#1}

\bibitem[{Brainard(2003)}]{brainard2003color}
David~H. Brainard. 2003.
\newblock Color appearance and color difference specification.
\newblock In \emph{The Science of Color}, 2 edition, pages 191--216. Elsevier, Oxford.

\bibitem[{Brochhagen and Boleda(2022)}]{brochhagen2022languages}
Thomas Brochhagen and Gemma Boleda. 2022.
\newblock When do languages use the same word for different meanings? the goldilocks principle in colexification.
\newblock \emph{Cognition}, 226:105179.

\bibitem[{Bréal(1897)}]{Bréal1897}
Michel Bréal. 1897.
\newblock \emph{Essai de sémantique: Science des significations}.
\newblock Librairie Hachette et Compagnie, Paris.

\bibitem[{Bybee(2010)}]{bybee2010language}
Joan Bybee. 2010.
\newblock \emph{Language, usage and cognition}.
\newblock Cambridge University Press.

\bibitem[{Campbell(2013)}]{campbell2013historical}
Lyle Campbell. 2013.
\newblock \emph{Historical linguistics}.
\newblock Edinburgh University Press.

\bibitem[{Carlsson et~al.(2024)Carlsson, Dubhashi, and Regier}]{carlsson2024cultural}
Emil Carlsson, Devdatt Dubhashi, and Terry Regier. 2024.
\newblock Cultural evolution via iterated learning and communication explains efficient color naming systems.
\newblock \emph{Journal of Language Evolution}, 9(1-2):49--66.

\bibitem[{Chaabouni et~al.(2020)Chaabouni, Kharitonov, Bouchacourt, Dupoux, and Baroni}]{chaabouni2020compositionality}
Rahma Chaabouni, Eugene Kharitonov, Diane Bouchacourt, Emmanuel Dupoux, and Marco Baroni. 2020.
\newblock Compositionality and generalization in emergent languages.
\newblock In \emph{Proceedings of the 58th Annual Meeting of the Association for Computational Linguistics}, pages 4427--4442.

\bibitem[{Chaabouni et~al.(2021)Chaabouni, Kharitonov, Dupoux, and Baroni}]{chaabouni2021communicating}
Rahma Chaabouni, Eugene Kharitonov, Emmanuel Dupoux, and Marco Baroni. 2021.
\newblock Communicating artificial neural networks develop efficient color-naming systems.
\newblock \emph{Proceedings of the National Academy of Sciences}, 118(12):e2016569118.

\bibitem[{Christiansen and Chater(2008)}]{christiansen2008language}
Morten~H Christiansen and Nick Chater. 2008.
\newblock Language as shaped by the brain.
\newblock \emph{Behavioral and brain sciences}, 31(5):489--509.

\bibitem[{De~Boer(2006)}]{de2006computer}
Bart De~Boer. 2006.
\newblock Computer modelling as a tool for understanding language evolution.
\newblock In \emph{Evolutionary epistemology, language and culture: A non-adaptationist, systems theoretical approach}, pages 381--406. Springer.

\bibitem[{Geeraerts(2020)}]{Geeraerts2020semantic-change}
Dirk Geeraerts. 2020.
\newblock \href {https://doi.org/10.1002/9781118788516.sem042} {\emph{Semantic Change}}, pages 1--24.
\newblock John Wiley \& Sons, Ltd.

\bibitem[{Goldberg(1995)}]{goldberg95}
Adele~E. Goldberg. 1995.
\newblock \emph{Construction grammar: a construction grammar approach to argument structure}.
\newblock University of Chicago Press.

\bibitem[{Goodman and Frank(2016)}]{goodman2016pragmatic}
Noah~D Goodman and Michael~C Frank. 2016.
\newblock Pragmatic language interpretation as probabilistic inference.
\newblock \emph{Trends in cognitive sciences}, 20(11):818--829.

\bibitem[{Graf et~al.(2016)Graf, Degen, Hawkins, and Goodman}]{graf2016animal}
Caroline Graf, Judith Degen, Robert~XD Hawkins, and Noah~D Goodman. 2016.
\newblock Animal, dog, or dalmatian? level of abstraction in nominal referring expressions.
\newblock In \emph{Proceedings of the Annual Meeting of the Cognitive Science Society}, volume~38.

\bibitem[{Gualdoni and Boleda(2024)}]{gualdoni-boleda-2024-objects}
Eleonora Gualdoni and Gemma Boleda. 2024.
\newblock \href {https://doi.org/10.18653/v1/2024.emnlp-main.1009} {Why do objects have many names? a study on word informativeness in language use and lexical systems}.
\newblock In \emph{Proceedings of the 2024 Conference on Empirical Methods in Natural Language Processing}, pages 18150--18163, Miami, Florida, USA. Association for Computational Linguistics.

\bibitem[{Gualdoni et~al.(2023)Gualdoni, Brochhagen, Mädebach, and Boleda}]{gualdoni_whats_2023}
Eleonora Gualdoni, Thomas Brochhagen, Andreas Mädebach, and Gemma Boleda. 2023.
\newblock What’s in a name? {A} large-scale computational study on how competition between names affects naming variation.
\newblock \emph{Journal of Memory and Language}, 133:104459.
\newblock Publisher: Elsevier.

\bibitem[{Gualdoni et~al.(2024)Gualdoni, Tucker, Levy, and Zaslavsky}]{gualdoni2024bridging}
Eleonora Gualdoni, Mycal Tucker, Roger Levy, and Noga Zaslavsky. 2024.
\newblock Bridging semantics and pragmatics in information-theoretic emergent communication.
\newblock \emph{Advances in Neural Information Processing Systems}, 37:21059--21078.

\bibitem[{Hawkins(2004)}]{hawkins2004efficiency}
John~A Hawkins. 2004.
\newblock \emph{Efficiency and complexity in grammars}.
\newblock OUP Oxford.

\bibitem[{Hawkins et~al.(2018)Hawkins, Franke, Smith, and Goodman}]{hawkins2018emerging}
Robert~XD Hawkins, Michael Franke, Kenny Smith, and Noah~D Goodman. 2018.
\newblock Emerging abstractions: Lexical conventions are shaped by communicative context.
\newblock In \emph{Proceedings of the Annual Meeting of the Cognitive Science Society}, volume~40.

\bibitem[{Hopper and Traugott(2003)}]{hopper2003grammaticalization}
Paul~J Hopper and Elizabeth~Closs Traugott. 2003.
\newblock \emph{Grammaticalization}.
\newblock Cambridge university press.

\bibitem[{Hurford(1989)}]{hurford1989biological}
James~R Hurford. 1989.
\newblock Biological evolution of the saussurean sign as a component of the language acquisition device.
\newblock \emph{Lingua}, 77:187--222.

\bibitem[{Kharitonov et~al.(2019)Kharitonov, Chaabouni, Bouchacourt, and Baroni}]{kharitonov-etal-2019-egg}
Eugene Kharitonov, Rahma Chaabouni, Diane Bouchacourt, and Marco Baroni. 2019.
\newblock \href {https://doi.org/10.18653/v1/D19-3010} {{EGG}: a toolkit for research on emergence of lan{G}uage in games}.
\newblock In \emph{Proceedings of the 2019 Conference on Empirical Methods in Natural Language Processing and the 9th International Joint Conference on Natural Language Processing (EMNLP-IJCNLP): System Demonstrations}, pages 55--60, Hong Kong, China. Association for Computational Linguistics.

\bibitem[{Kharitonov et~al.(2020)Kharitonov, Chaabouni, Bouchacourt, and Baroni}]{kharitonov2020entropy}
Eugene Kharitonov, Rahma Chaabouni, Diane Bouchacourt, and Marco Baroni. 2020.
\newblock Entropy minimization in emergent languages.
\newblock In \emph{International Conference on Machine Learning}, pages 5220--5230. PMLR.

\bibitem[{Kingma and Ba(2015)}]{kingma2015adam}
Diederik~P. Kingma and Jimmy Ba. 2015.
\newblock \href {https://arxiv.org/abs/1412.6980} {Adam: A method for stochastic optimization}.
\newblock In \emph{The 3rd International Conference on Learning Representations (ICLR)}, San Diego, CA, USA.

\bibitem[{Kobrock et~al.(2025)Kobrock, Ohmer, Bruni, and Gotzner}]{kobrock2025agents}
Kristina Kobrock, Xenia Ohmer, Elia Bruni, and Nicole Gotzner. 2025.
\newblock Agents generalize to novel levels of abstraction by using adaptive linguistic strategies.

\bibitem[{Kobrock et~al.(2024)Kobrock, Ohmer, Bruni, and Gotzner}]{kobrock-etal-2024-context}
Kristina Kobrock, Xenia~Isabel Ohmer, Elia Bruni, and Nicole Gotzner. 2024.
\newblock \href {https://aclanthology.org/2024.lrec-main.339/} {Context shapes emergent communication about concepts at different levels of abstraction}.
\newblock In \emph{Proceedings of the 2024 Joint International Conference on Computational Linguistics, Language Resources and Evaluation (LREC-COLING 2024)}, pages 3831--3848, Torino, Italia. ELRA and ICCL.

\bibitem[{Labov(2010)}]{labov2010principles}
William Labov. 2010.
\newblock \emph{Principles of linguistic change, Volume 3: Cognitive and cultural factors}, volume~3.
\newblock John Wiley \& Sons.

\bibitem[{Lazaridou and Baroni(2020)}]{lazaridou2020emergent}
Angeliki Lazaridou and Marco Baroni. 2020.
\newblock Emergent multi-agent communication in the deep learning era.
\newblock \emph{arXiv preprint arXiv:2006.02419}.

\bibitem[{Lazaridou et~al.(2018)Lazaridou, Hermann, Tuyls, and Clark}]{lazaridou2018emergence}
Angeliki Lazaridou, Karl~Moritz Hermann, Karl Tuyls, and Stephen Clark. 2018.
\newblock \href {https://openreview.net/forum?id=HJGv1Z-AW} {Emergence of linguistic communication from referential games with symbolic and pixel input}.
\newblock In \emph{International Conference on Learning Representations}.

\bibitem[{Lian et~al.(2023)Lian, Bisazza, and Verhoef}]{lian2023communication}
Yuchen Lian, Arianna Bisazza, and Tessa Verhoef. 2023.
\newblock \href {https://doi.org/10.1162/tacl_a_00587} {{Communication Drives the Emergence of Language Universals in Neural Agents: Evidence from the Word-order/Case-marking Trade-off}}.
\newblock \emph{Transactions of the Association for Computational Linguistics}, 11:1033--1047.

\bibitem[{Lian et~al.(2025)Lian, Bisazza, and Verhoef}]{lian2025simulating}
Yuchen Lian, Arianna Bisazza, and Tessa Verhoef. 2025.
\newblock \href {https://doi.org/10.48550/arXiv.2502.04038} {Simulating the emergence of differential case marking with communicating neural-network agents}.
\newblock In \emph{Proceedings of the 47th Annual Conference of the Cognitive Science Society (CogSci)}.

\bibitem[{Lian et~al.(2024)Lian, Verhoef, and Bisazza}]{lian-etal-2024-nellcom}
Yuchen Lian, Tessa Verhoef, and Arianna Bisazza. 2024.
\newblock \href {https://doi.org/10.18653/v1/2024.conll-1.19} {{N}e{LLC}om-{X}: A comprehensive neural-agent framework to simulate language learning and group communication}.
\newblock In \emph{Proceedings of the 28th Conference on Computational Natural Language Learning}, pages 243--258, Miami, FL, USA. Association for Computational Linguistics.

\bibitem[{Michel et~al.(2023)Michel, Rita, Mathewson, Tieleman, and Lazaridou}]{michel2023revisiting}
Paul Michel, Mathieu Rita, Kory~Wallace Mathewson, Olivier Tieleman, and Angeliki Lazaridou. 2023.
\newblock \href {https://openreview.net/forum?id=n-UHRIdPju} {Revisiting populations in multi-agent communication}.
\newblock In \emph{The Eleventh International Conference on Learning Representations}.

\bibitem[{Monroe et~al.(2017)Monroe, Hawkins, Goodman, and Potts}]{monroe-etal-2017-colors}
Will Monroe, Robert~X.D. Hawkins, Noah~D. Goodman, and Christopher Potts. 2017.
\newblock \href {https://doi.org/10.1162/tacl_a_00064} {Colors in context: A pragmatic neural model for grounded language understanding}.
\newblock \emph{Transactions of the Association for Computational Linguistics}, 5:325--338.

\bibitem[{Paszke et~al.(2017)Paszke, Gross, Chintala, Chanan, Yang, DeVito, Lin, Desmaison, Antiga, and Lerer}]{paszke2017automatic}
Adam Paszke, Sam Gross, Soumith Chintala, Gregory Chanan, Edward Yang, Zachary DeVito, Zeming Lin, Alban Desmaison, Luca Antiga, and Adam Lerer. 2017.
\newblock \href {https://openreview.net/forum?id=BJJsrmfCZ} {Automatic differentiation in pytorch}.
\newblock In \emph{NIPS 2017 Workshop on Autodiff}.

\bibitem[{Regier et~al.(2015)Regier, Kemp, and Kay}]{regier2015word}
Terry Regier, Charles Kemp, and Paul Kay. 2015.
\newblock Word meanings across languages support efficient communication.
\newblock \emph{The handbook of language emergence}, pages 237--263.

\bibitem[{Sedivy(2003)}]{sedivy2003pragmatic}
Julie~C Sedivy. 2003.
\newblock Pragmatic versus form-based accounts of referential contrast: Evidence for effects of informativity expectations.
\newblock \emph{Journal of psycholinguistic research}, 32:3--23.

\bibitem[{Sedivy et~al.(1999)Sedivy, Tanenhaus, Chambers, and Carlson}]{sedivy1999achieving}
Julie~C Sedivy, Michael~K Tanenhaus, Craig~G Chambers, and Gregory~N Carlson. 1999.
\newblock Achieving incremental semantic interpretation through contextual representation.
\newblock \emph{Cognition}, 71(2):109--147.

\bibitem[{Sharma et~al.(2005)Sharma, Wu, and Dalal}]{sharma2005ciede2000}
Gaurav Sharma, Wencheng Wu, and Edul~N Dalal. 2005.
\newblock The ciede2000 color-difference formula: Implementation notes, supplementary test data, and mathematical observations.
\newblock \emph{Color Research \& Application}, 30(1):21--30.

\bibitem[{Smith(2004)}]{smith2004evolution}
Kenny Smith. 2004.
\newblock The evolution of vocabulary.
\newblock \emph{Journal of theoretical biology}, 228(1):127--142.

\bibitem[{Smith(2022)}]{smith2022language}
Kenny Smith. 2022.
\newblock How language learning and language use create linguistic structure.
\newblock \emph{Current Directions in Psychological Science}, 31(2):177--186.

\bibitem[{Steels(1997)}]{steels1997synthetic}
Luc Steels. 1997.
\newblock The synthetic modeling of language origins.
\newblock \emph{Evolution of communication}, 1(1):1--34.

\bibitem[{Steels(2005)}]{steels2005emergence}
Luc Steels. 2005.
\newblock The emergence and evolution of linguistic structure: from lexical to grammatical communication systems.
\newblock \emph{Connection science}, 17(3-4):213--230.

\bibitem[{White et~al.(2020)White, Mu, and Goodman}]{white2020learning}
Julia White, Jesse Mu, and Noah~D Goodman. 2020.
\newblock Learning to refer informatively by amortizing pragmatic reasoning.
\newblock In \emph{Proceedings of the Annual Meeting of the Cognitive Science Society}, volume~42.

\bibitem[{Williams(1992)}]{williams1992simple}
Ronald~J Williams. 1992.
\newblock Simple statistical gradient-following algorithms for connectionist reinforcement learning.
\newblock \emph{Machine learning}, 8:229--256.

\bibitem[{Winters et~al.(2015)Winters, Kirby, and Smith}]{winters2015languages}
James Winters, Simon Kirby, and Kenny Smith. 2015.
\newblock Languages adapt to their contextual niche.
\newblock \emph{Language and Cognition}, 7(3):415--449.

\bibitem[{Zarrie{\ss} and Schlangen(2019)}]{zarriess-schlangen-2019-know}
Sina Zarrie{\ss} and David Schlangen. 2019.
\newblock \href {https://doi.org/10.18653/v1/P19-1063} {Know what you don`t know: Modeling a pragmatic speaker that refers to objects of unknown categories}.
\newblock In \emph{Proceedings of the 57th Annual Meeting of the Association for Computational Linguistics}, pages 654--659, Florence, Italy. Association for Computational Linguistics.

\bibitem[{Zaslavsky et~al.(2018)Zaslavsky, Kemp, Regier, and Tishby}]{zaslavsky2018efficient}
Noga Zaslavsky, Charles Kemp, Terry Regier, and Naftali Tishby. 2018.
\newblock Efficient compression in color naming and its evolution.
\newblock \emph{Proceedings of the National Academy of Sciences}, 115(31):7937--7942.

\bibitem[{Zhang et~al.(2024{\natexlab{a}})Zhang, Verhoef, van Noord, and Bisazza}]{zhang-etal-2024-endowing}
Yuqing Zhang, Tessa Verhoef, Gertjan van Noord, and Arianna Bisazza. 2024{\natexlab{a}}.
\newblock \href {https://aclanthology.org/2024.lrec-main.516/} {Endowing neural language learners with human-like biases: A case study on dependency length minimization}.
\newblock In \emph{Proceedings of the 2024 Joint International Conference on Computational Linguistics, Language Resources and Evaluation (LREC-COLING 2024)}, pages 5819--5832, Torino, Italia. ELRA and ICCL.

\bibitem[{Zhang et~al.(2024{\natexlab{b}})Zhang, Verhoef, van Noord, and Bisazza}]{zhang2024neural}
Yuqing Zhang, Tessa Verhoef, Gertjan van Noord, and Arianna Bisazza. 2024{\natexlab{b}}.
\newblock Neural-agent language learning and communication: Emergence of dependency length minimization.
\newblock In \emph{Proceedings of the Annual Meeting of the Cognitive Science Society}, volume~46.

\end{thebibliography}

\appendix

\section{Datasets and model training}

\label{app:appendix_data_model}
Colors in the dataset are represented as a three-dimensional vector in CIELAB color space. CIELAB was chosen for its perceptual uniformity, where Euclidean distances approximate color differences perceived by the human eye more accurately than in other color models \citep{brainard2003color}.

\paragraph{Data generation}
For RL training and evaluation, we generate a large number of color triplets for the three context types (far, split, close), following the sampling procedure described by \citet{monroe-etal-2017-colors}. The conditions are defined as follows: (1) close, where all three colors are within a perceptible threshold distance $\theta$ from one another; (2) split, where one distractor is within a distance of $\theta$ from the target while the other is farther away; and (3) far, where all colors are farther than $\theta$ from each other. Colors are rejection-sampled uniformly from the RGB space to satisfy these constraints. 

To ensure that all sampled color differences are perceptible to human vision, we used the most recent CIEDE2000 standard to measure color distances \citep{sharma2005ciede2000}, which is calibrated to human perceptual judgments. All distances were constrained to exceed a lower bound of 5 (just noticeable difference), and we used a threshold value of $\theta = 20$ to define the context types.


\paragraph{Model training}
Hyperparameters were selected based on preliminary experiments. Speakers and listeners both use feedforward neural network (FNN) modules with a hidden dimension of 512. Each FNN module consists of a linear layer followed by batch normalization, a ReLU activation, and a dropout layer. Both SL and RL phases use the default Adam optimizer \citep{kingma2015adam} implemented in PyTorch \citep{paszke2017automatic}, with a learning rate of $1 \times 10^{-5}$ and a batch size of 32. Each training phase runs for 30 epochs, and all experiments are repeated across 10 random seeds. For the RL phase, we set the entropy regularization parameter as 0.15 for both the speaker and the listener model.  

The implementation is partly based on the EGG toolkit \citep{kharitonov-etal-2019-egg}.

\section{Pairwise comparisons}
\label{app:comparison_parta}

\subsection{Pairwise comparisons for different SL and RL training pipelines}
To evaluate how the relationship between context ease and word informativeness varies across training pipelines, we fit two linear mixed-effects models. The first model includes an interaction term between context ease and training pipeline, allowing the slope of context ease to vary by pipeline. The second model excludes this interaction, assuming a uniform effect of context ease across all pipelines. A likelihood ratio test comparing the two models confirmed that including the interaction significantly improves model fit ($\chi^2(4) = 58.70$, $p < .001$), indicating that the effect of context ease on informativeness differs by training pipeline. Each model includes random intercepts for seed and target chip, accounting for variation across model initializations and color items.

Table \ref{tab:pairwise_context_condition} shows pairwise comparisons for the interaction between context ease and different training pipelines. 
The negative estimates indicate that certain training pipelines lead to a stronger decrease in informativeness as context ease increases. In other words, agents trained under those pipelines are more sensitive to context difficulty when naming targets. Specifically, agents trained without access to context throughout (SL$-$, SL$-$RL$-$) show little to no sensitivity to context ease, while pipelines involving context access, either during SL or RL, lead to significantly steeper declines in informativeness with increasing context ease (e.g., SL$-$RL$+$ vs.\ SL$-$: $\beta = -0.001$, $p < 0.001$; SL$+$ vs.\ SL$-$: $\beta = -0.001$, $p < 0.001$). However, the differences among pipelines that include context (SL$-$RL$+$, SL$+$, SL$+$RL$+$) are not statistically significant, suggesting that any context exposure is sufficient to promote pragmatic adaptation.

\begin{table}[h!]
\centering \small
\resizebox{\linewidth}{!}{
\begin{tabular}{lrr}
\toprule
\textbf{Comparison} & \textbf{Estimate} & \textbf{p-value} \\
\midrule
SL$-$ $\to$ SL$-$RL$-$      & -0.000 & n.s. \\
SL$-$ $\to$ SL$-$RL$+$      & -0.001 & $p < 0.001$ \\
SL$-$ $\to$ SL$+$         & -0.001 & $p < 0.001$ \\
SL$-$ $\to$ SL$+$RL$+$    & -0.001 & $p < 0.001$ \\
SL$-$RL$-$ $\to$ SL$-$RL$+$   & -0.001 & $p < 0.001$ \\
SL$-$RL$-$ $\to$ SL$+$      & -0.001 & $p < 0.01$ \\
SL$-$RL$-$ $\to$ SL$+$RL$+$ & -0.001 & $p < 0.01$ \\
SL$-$RL$+$ $\to$ SL$+$      & 0.000 & n.s. \\
SL$-$RL$+$ $\to$ SL$+$RL$+$ & 0.000 & n.s. \\
SL$+$ $\to$ SL$+$RL$+$    & -0.000 & n.s. \\
\bottomrule
\end{tabular}}
\caption{Pairwise comparisons for the interaction between context ease and different training pipelines. Estimates are rounded to three decimals. The “n.s.” label indicates not significant. For instance, the comparison SL$-$~$\to$~SL$-$RL$+$ shows an estimated slope difference of $-0.001$ with $p < 0.001$. 
This indicates that the learning curve in the SL$-$RL$+$ condition is significantly steeper than in the SL$-$ condition, 
although the absolute difference in slope is very small.}
\label{tab:pairwise_context_condition}
\end{table}

\subsection{Pairwise comparisons for varying context distributions}
\noindent Similarly, to examine how training under varying context distributions during RL affects agents' sensitivity to context ease, we fit two linear mixed-effects models. The first model includes an interaction between context ease and RL training condition, allowing the slope of informativeness as a function of context ease to vary across conditions. The second model excludes this interaction, assuming a uniform slope across all training conditions. A likelihood ratio test confirmed that the interaction significantly improves model fit ($\chi^2(2) = 174.35$, $p < .001$), indicating that the effect of context ease on informativeness differs depending on the distribution of contexts during RL training.
Table \ref{tab:pairwise_comparisons_b} shows pairwise comparisons for the interaction between context ease and varying context distributions during RL training.
The slope is steepest in the AllClose condition, significantly more negative than in AllFar ($\beta = -0.002$, $p < .001$) and HalfHalf ($\beta = -0.001$, $p < .001$). The HalfHalf condition is also significantly steeper than AllFar ($\beta = -0.001$, $p < .001$).\footnote{Here, $\beta$ denotes the estimated difference in slopes between conditions.} 
Specifically, agents trained exclusively on difficult contexts learn to produce more informative words and show slightly stronger context sensitivity. In contrast, agents trained on easy contexts show a weaker adaptation effect, less context sensitivity, likely reflecting their reduced need to modulate informativeness by context ease.

\begin{table}[ht]
\centering \small
\begin{tabular}{ccc}
\toprule
\textbf{Comparison} & \textbf{Estimate} & \textbf{$p$-value} \\
\midrule
AllFar $\to$ AllClose       & -0.002  & p$<$0.001 \\
AllFar $\to$ HalfHalf       & -0.001  & p$<$0.001 \\
AllClose $\to$ HalfHalf     & 0.001 & p$<$0.001 \\
\bottomrule
\end{tabular}
\caption{Pairwise comparisons for the interaction between context ease and varying context distributions during RL training.}
\label{tab:pairwise_comparisons_b}
\end{table}

\section{Relationship between context ease and word informativeness}
\label{app:informativess_scatter_parta}

\begin{figure*}[t]
  \centering
  \includegraphics[width=\linewidth]{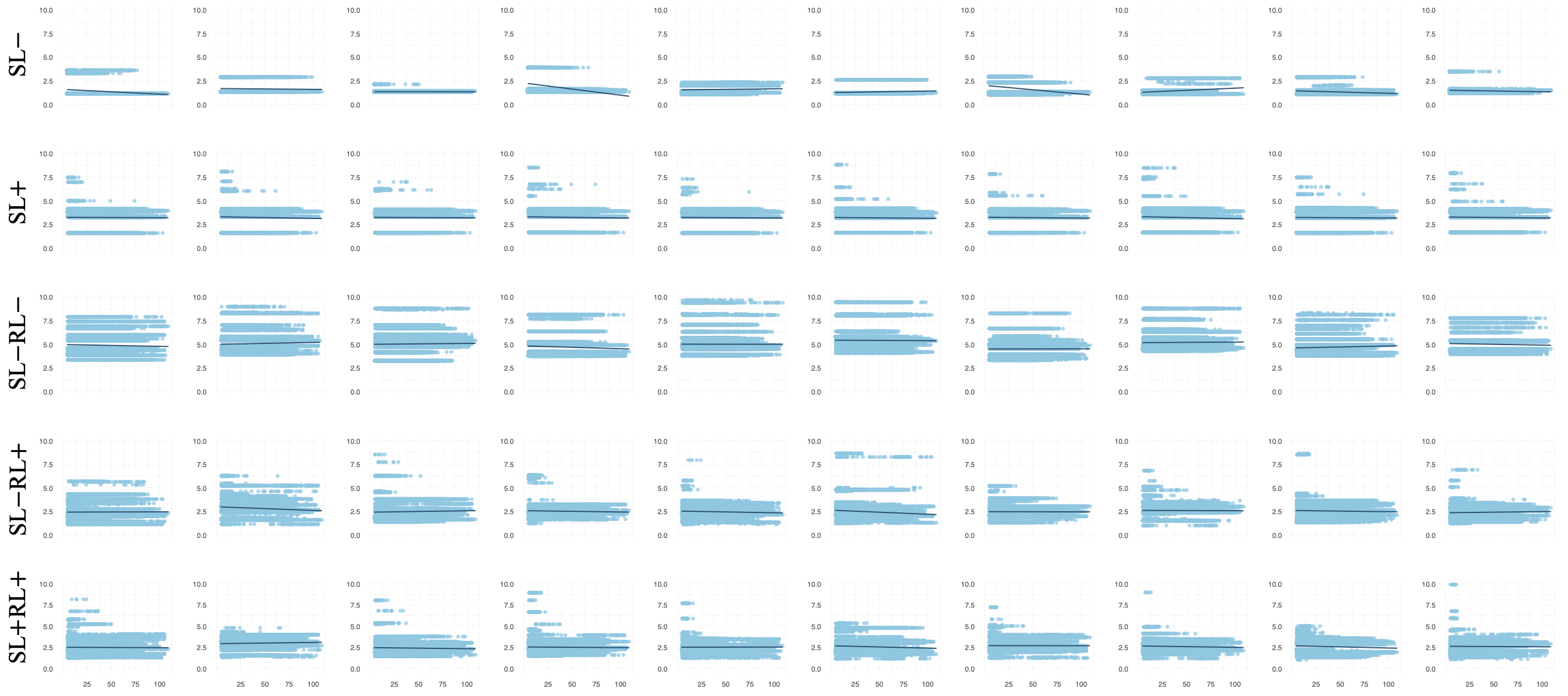}
  \caption{Word informativeness as a function of context ease on \textsc{test}$_{gen,distH}$ for all seeds and for different SL and RL training pipelines.}
  \label{fig:informativeness_plot_all_a}
\end{figure*}

\begin{figure*}[h]
  \centering
  \includegraphics[width=\linewidth]{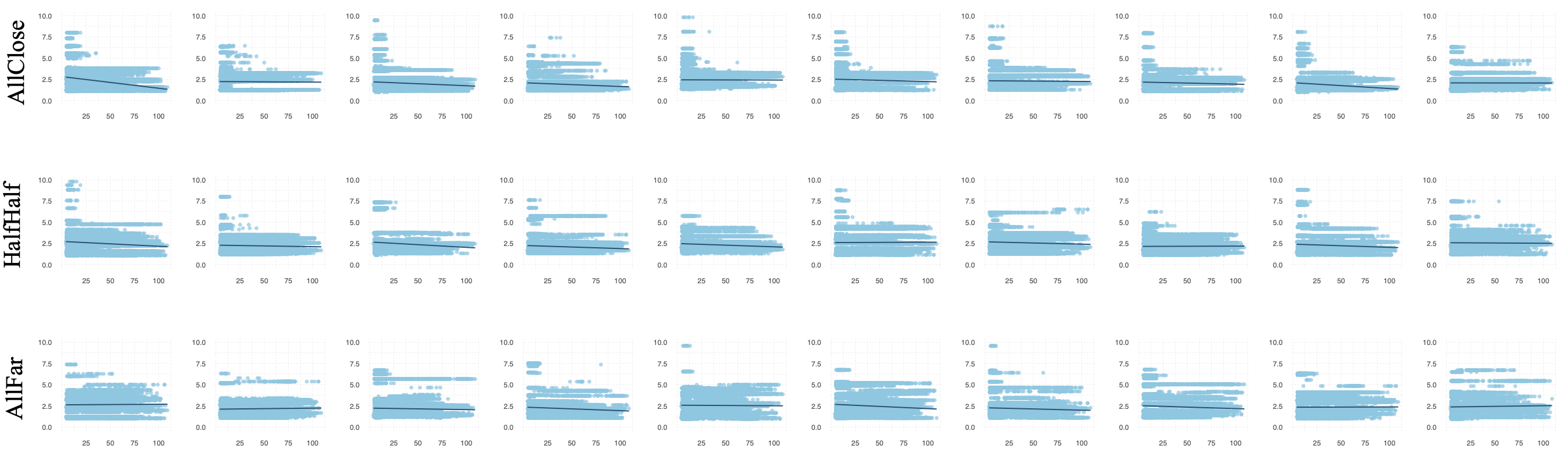}
  \caption{Word informativeness as a function of context ease on \textsc{test}$_{gen,dist50}$ for all seeds and for varying context distributions during RL training.}
  \label{fig:informativeness_plot_b}
\end{figure*}

\subsection{Visualization for different SL and RL training pipelines}

Figure~\ref{fig:informativeness_plot_all_a} shows the relationship between context ease and word informativeness across different SL and RL training pipelines and all random seeds.

While there is some variation across agents within each condition, trends still emerge. Agents trained without access to context, either after SL (SL$-$) or after both SL and RL (SL$-$RL$-$), show no sensitivity to $E_{ctx}$. In contrast, agents that were exposed to context during SL (SL$+$), RL (SL$-$RL$+$), or both phases (SL$+$RL$+$), tend to produce more pragmatic language. In these settings, word informativeness decreases slightly but significantly as the context becomes easier, indicating that context-aware agents adapt their lexical choices to the communicative needs of the context.

\subsection{Visualization for varying RL context distributions}

Figure~\ref{fig:informativeness_plot_b} shows the relationship between context ease and word informativeness across different SL and RL training pipelines and all random seeds. 
Agents trained exclusively on difficult contexts (AllClose) produce more informative words in harder contexts and show slightly stronger context sensitivity compared to other conditions.

\section{Entropy of informativeness over word types}
\label{app:entropy_informativeness}
To examine how informativeness is distributed across the color terms used by agents, we computed the entropy of word informativeness over word types for each condition and after each training phase. First, we normalized the informativeness scores by converting them into a probability distribution: each word's informativeness score was divided by the total informativeness summed over all word types. We calculated the following equation:
\begin{equation}
P(w_i) = \frac{I(w_i)}{\sum_{j=1}^{N} I(w_j)}
\end{equation}
where $P(w_i)$ is the normalized informativeness score, with other words the probability of the informativeness score. After obtaining the normalized score, we computed Shannon entropy over this distribution to assess how evenly informativeness is spread across the vocabulary with the following formula.
\begin{equation}
H = - \sum_{i=1}^{N} P(w_i) \log P(w_i)
\end{equation}
where \emph{H} refers to Shannon's entropy.

\subsection{Entropy of informativeness for different SL and RL training pipelines}

\begin{figure}[h!]
    \centering
    \includegraphics[width=0.75\linewidth]{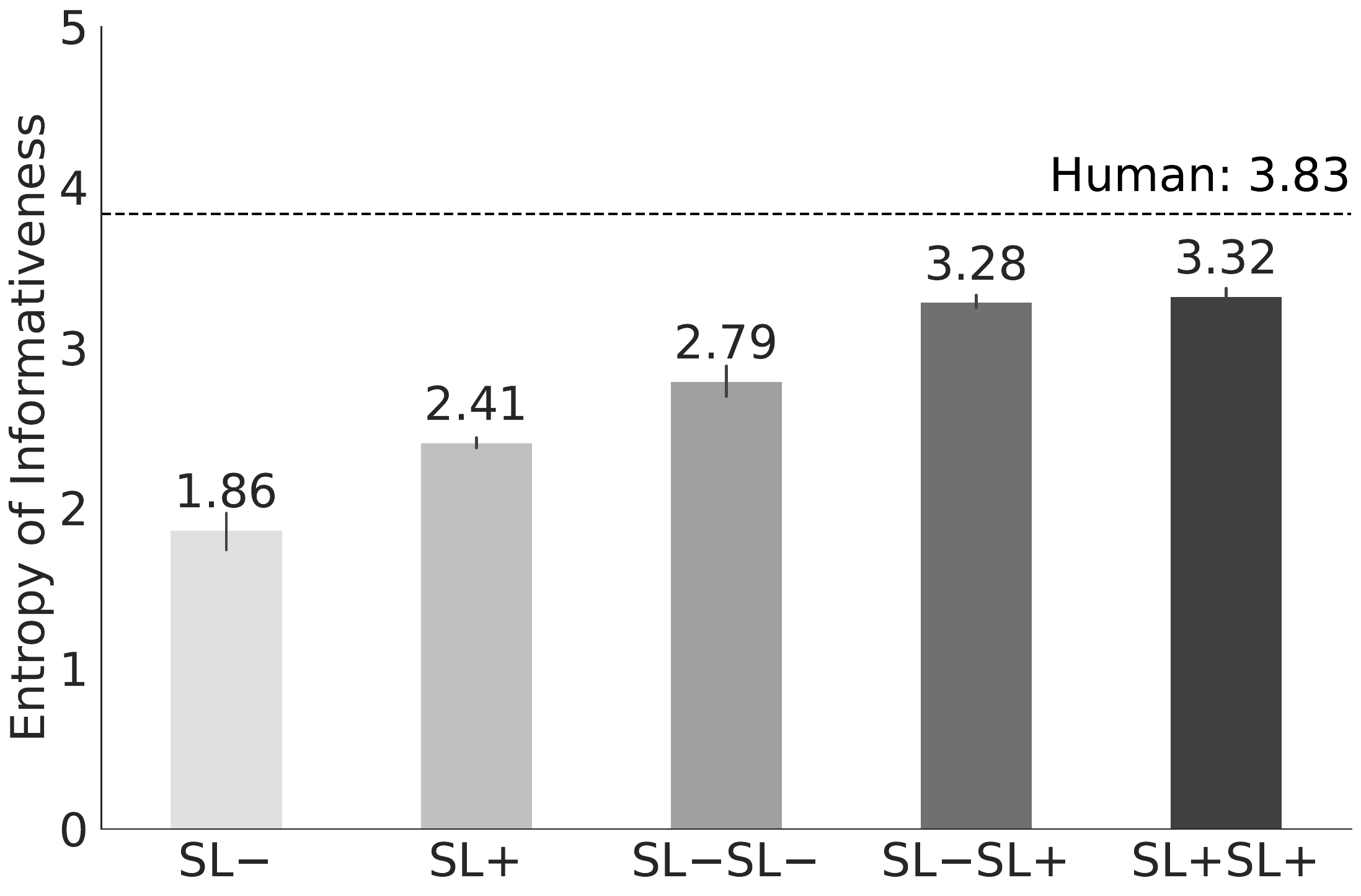} 
    \caption{Entropy of informativeness over word types under different SL and RL training pipelines. Only entropy computed on the full test set is reported, as the distribution of far and close contexts is highly imbalanced, with far contexts being more frequent.}
    \label{fig:entropy_informativeness_a}
\end{figure}

This section analyzes the entropy of informativeness across different training pipelines that combine SL and RL. Figure \ref{fig:entropy_informativeness_a} illustrates the entropy values for different SL and RL training conditions.

The condition SL$+$RL$-$ shows a higher entropy, indicating a more balanced distribution of informativeness across a broader vocabulary. This suggests that the model has access to a richer set of words, potentially improving its ability to adapt to diverse contexts and enhancing communication efficiency. In contrast, SL$-$RL$-$ exhibits lower entropy, with fewer words carrying most of the informativeness. This could imply that the model is over-relying on a small subset of words, making it more prone to overfitting and limiting its generalization ability, particularly in new or varied contexts. Therefore, the SL$+$RL$+$ condition may facilitate better adaptability and clearer communication, while the SL$-$RL$-$ condition risks reduced flexibility.
Moreover, the entropy values for the conditions after RL are considerably higher. This indicates that the model, having been trained with both context-aware SL and RL, is most capable of adapting to varied contexts and communicating more effectively.

\subsection{Entropy of informativeness for varying context distributions} 

\begin{figure}[h!]
    \centering
    \includegraphics[width=\linewidth]{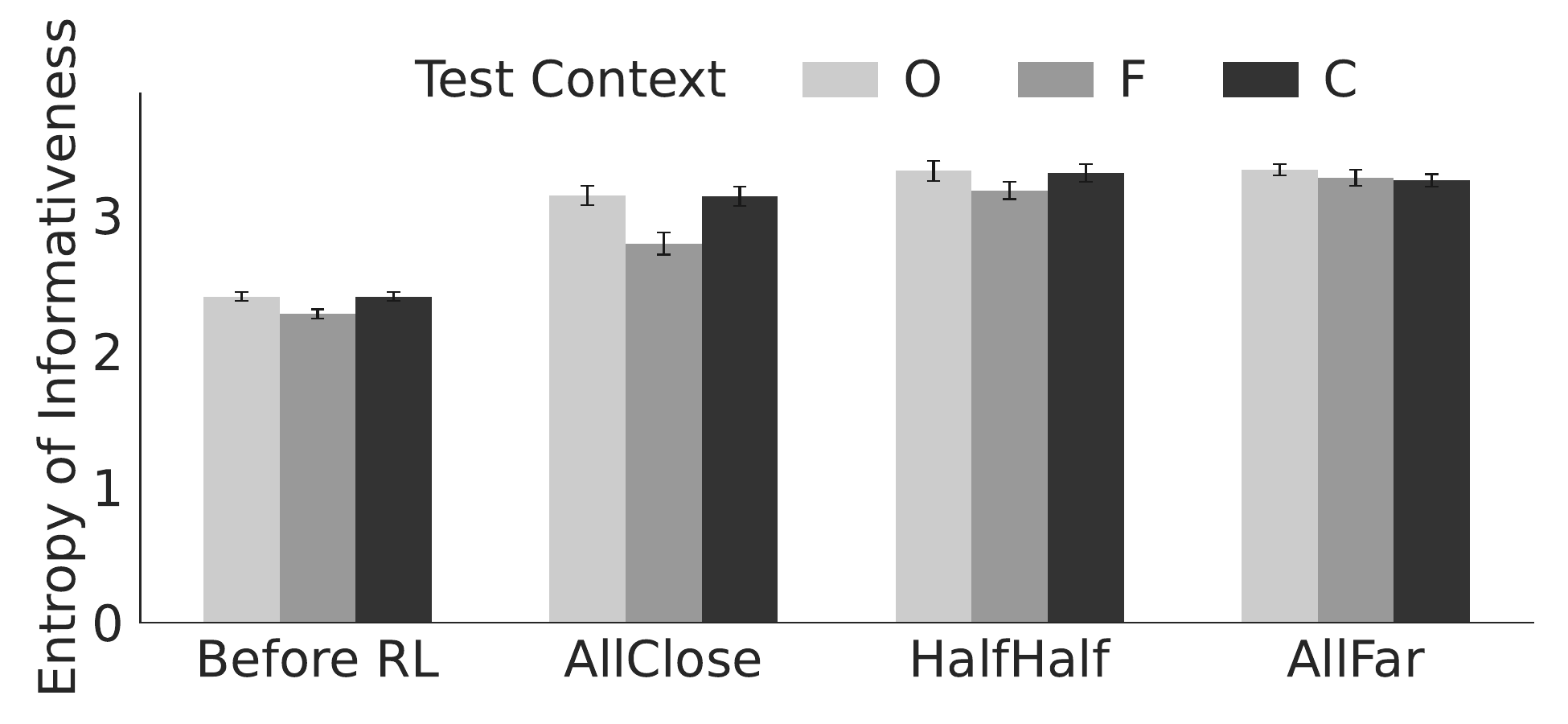}  
    \caption{Entropy of informativeness over word types after agents being trained on varying context distributions during RL (O: overall; F: far; C: close).}
    \label{fig:entropy_informativeness_b}
\end{figure}

Figure \ref{fig:entropy_informativeness_b} illustrates the entropy of informativeness for varying context distributions during RL training. The entropy of informativeness is higher for the close context compared to the far context in each condition. However, when the proportion of far context increases in the training data, the difference in entropy between far and close contexts decreases. This suggests that a higher presence of far context in the training data reduces the diversity of informativeness.

\section{Lexical diversity}
\label{app:lexical_diversity_parta}
\begin{figure}[h!]
  \centering
  \includegraphics[width=\linewidth]{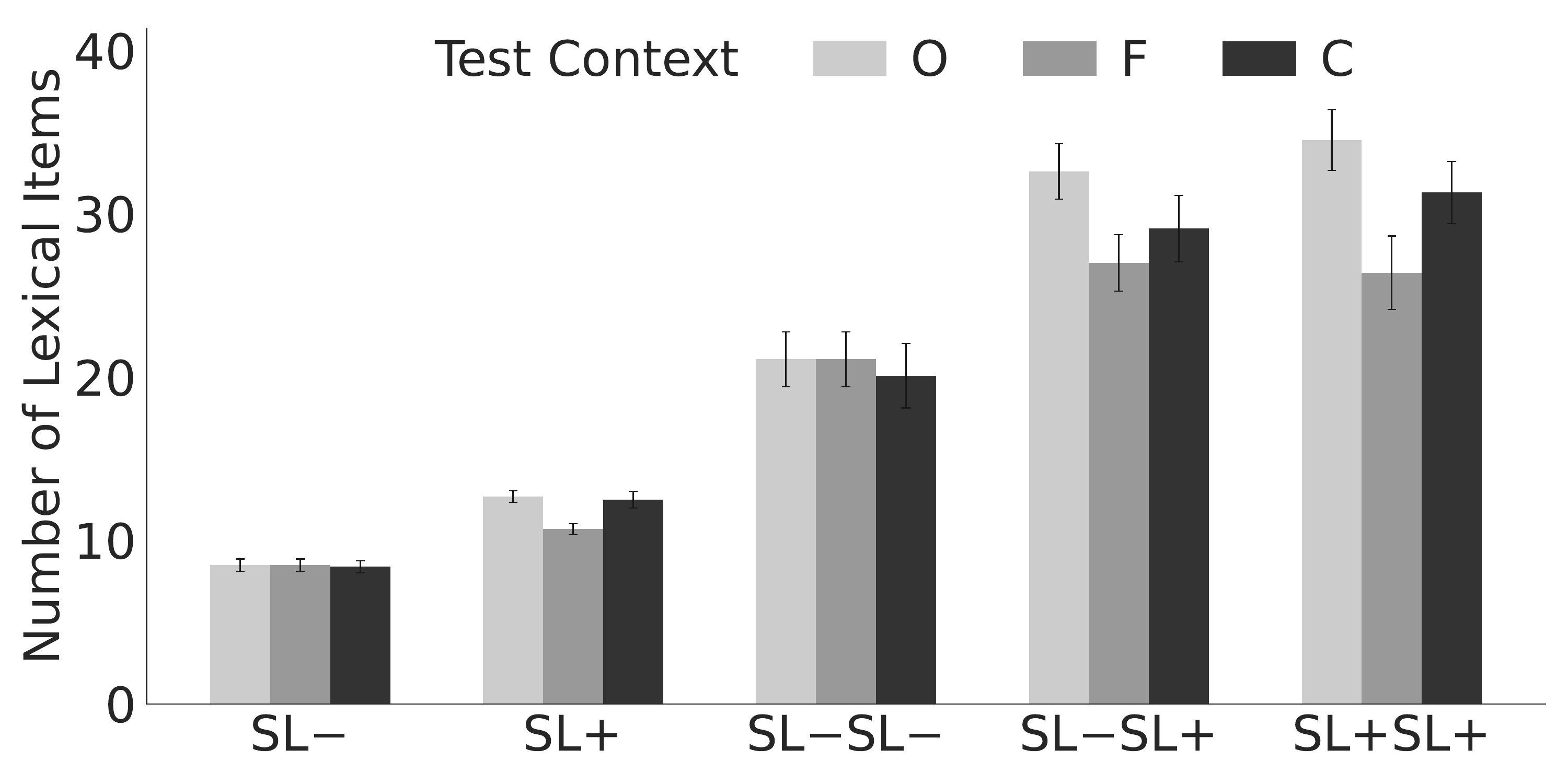}
  \caption{Number of distinct lexical items used before RL and after RL training. Error bars represent 95\% confidence intervals computed across the test set (O: overall; F: far; C: close).}
  \label{fig:word_type_usage_parta}
\end{figure}

Although the distribution of far and close contexts in \textsc{test}$_{gen,distH}$ is imbalanced, with the close condition represented by a smaller number of examples, distinct patterns of lexical diversity still emerge (see Figure \ref{fig:word_type_usage_parta}).

RL training substantially leads to higher lexical diversity, especially in the SL$+$RL$+$ condition. Moreover, in context-aware settings (SL$+$, SL$-$RL$+$, SL$+$RL$+$), agents tend to produce a greater variety of word types in the close test set compared to the far test set, despite the former being smaller in size.

These findings suggest that access to contextual information not only promotes pragmatic language use but also supports the development of a more flexible and expressive lexicon, especially in more challenging communicative settings.

\section{System-level informativeness}
\label{app:system_informativeness_partb}

\begin{figure}[h]
  \centering
  \includegraphics[width=\linewidth]{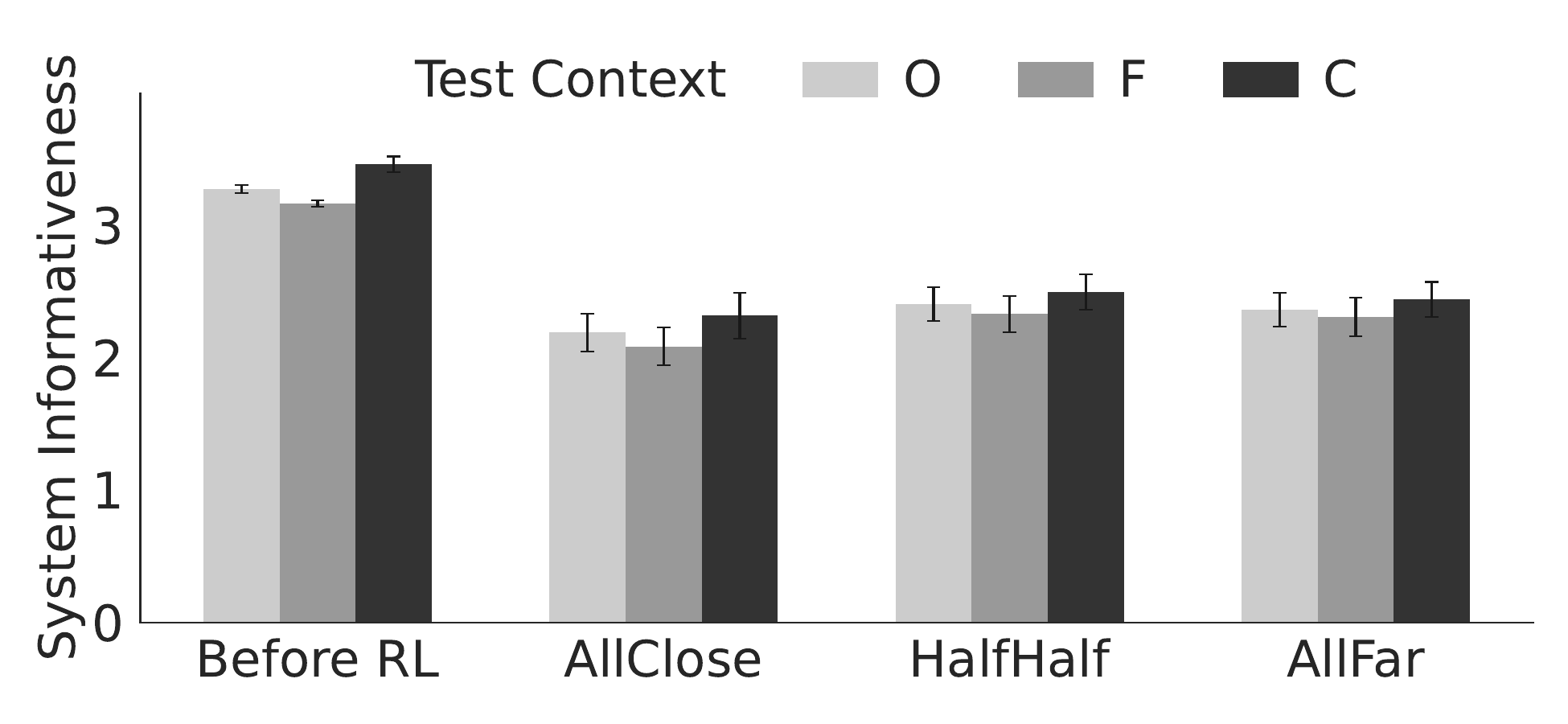} 
  \caption{System-level informativeness before RL and after RL training on varying context conditions (O: overall; F: far; C: close).}
  \label{fig:system_informativeness_partb}
\end{figure}

Figure \ref{fig:system_informativeness_partb} shows the system-level informativeness for varying context distributions before RL and after RL training on varying context conditions.

Before RL, the system-level informativeness is higher in the close context compared to the far context. This difference progressively diminishes as the context distribution shifts from AllClose to HalfHalf and then to AllFar. In the AllFar condition, the informativeness gap nearly disappears, indicating that increased exposure to far contexts encourages the agents to develop a more uniform lexicon that generalizes consistently across different context types. 

Notably, the system-level informativeness of neural agents before RL closely matches that observed in human production, both of which are higher than the informativeness after RL. This suggests that SL causes agents to replicate the distribution found in the human training data, while communication learning drives agents to become more efficient by reducing system-level informativeness.

\section{Visualization of word embeddings}

\label{app:word_embeddings}

\begin{figure}[t]
  \centering
    \includegraphics[width=1.1\linewidth]{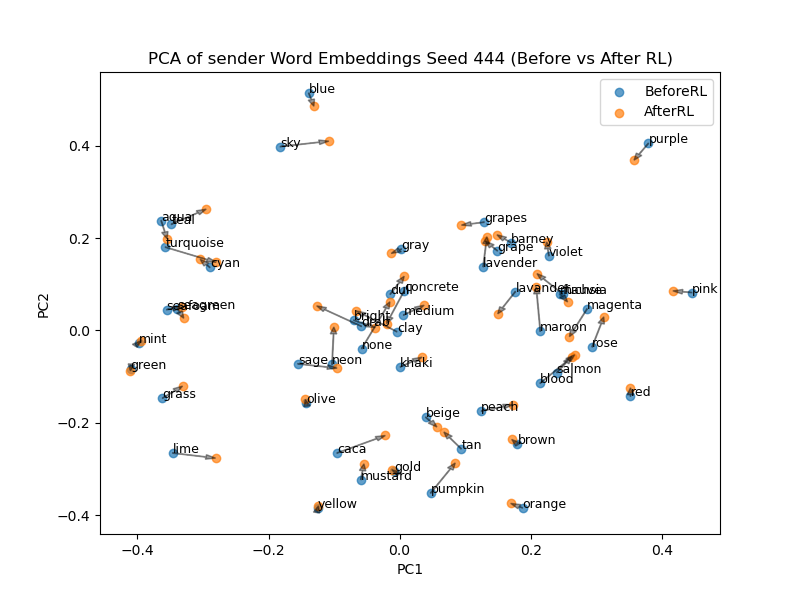}
  \caption{PCA projection of the speaker embeddings of color terms, with positions before and after RL, for a representative seed of the SL+RL+ condition.}
\label{fig:word_embeddings}
\end{figure}

The lexicon-related measures we use are based on word denotations (i.e. on the speaking agents' productions), allowing us to compare agents to humans. However, for agents, we can also inspect their internal lexical representations, in the form of word embeddings.
Figure~\ref{fig:word_embeddings} shows a 2-D projection of the output word embeddings (unembedding matrix) for a representative speaking agent trained by SL$+$RL$+$ in HalfHalf condition.
We can see that the embeddings reasonably reflect color-to-color relations:
on the left part, we go from blue to green as we move from top to bottom; in the middle, we find yellows and browns; on the right, the different reds. Aqua, turquoise and cyan are near to each other and between blue and green; etc.

During RL, the representations of many words change noticeably, but the overall semantic relationships and main color clusters are largely preserved.
We leave to future work a more in-depth analysis of how the agents' word embedding spaces change through interaction, and how those changes relate to the semantic drift observed in the agents' production.

\section{Color denotation in visual space} 

\begin{figure}[h]
  \centering
  \begin{subfigure}[b]{0.32\linewidth}
    \centering
    \includegraphics[width=\linewidth]{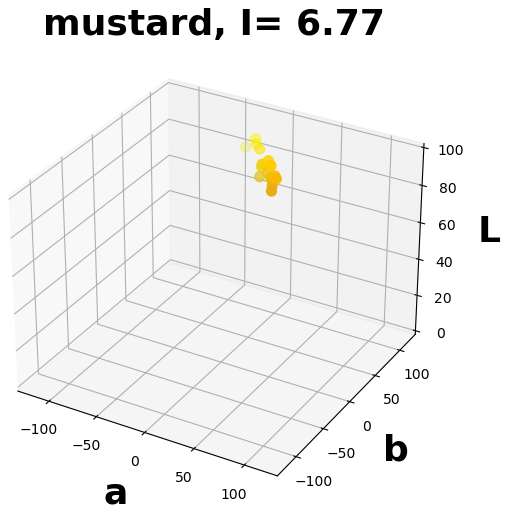}
  \end{subfigure}
  \begin{subfigure}[b]{0.32\linewidth}
    \centering
    \includegraphics[width=\linewidth]{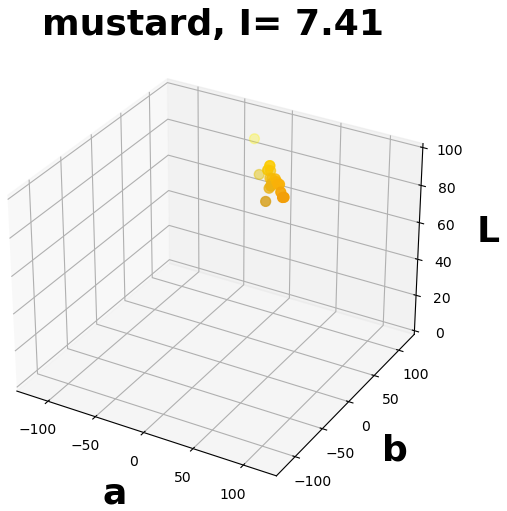}
  \end{subfigure}
  \begin{subfigure}[b]{0.32\linewidth}
    \centering
    \includegraphics[width=\linewidth]{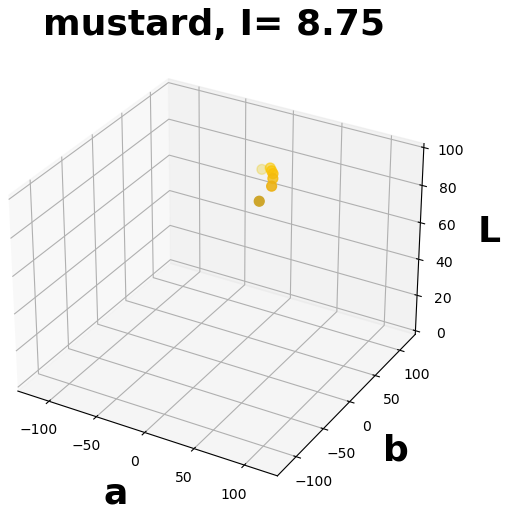}
  \end{subfigure}
    \vspace{1em}

  \begin{subfigure}[b]{0.32\linewidth}
    \centering
    \includegraphics[width=\linewidth]{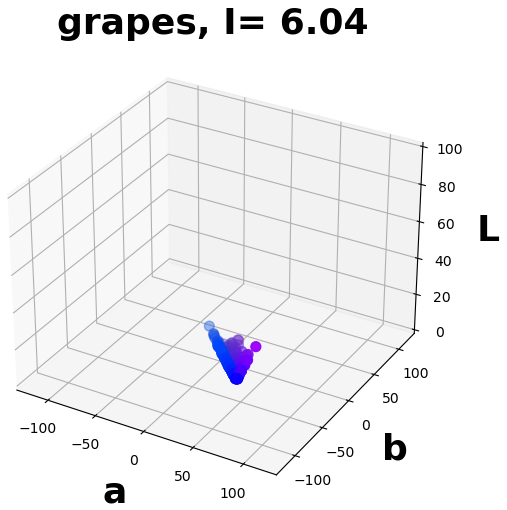}
  \end{subfigure}
  \begin{subfigure}[b]{0.32\linewidth}
    \centering
    \includegraphics[width=\linewidth]{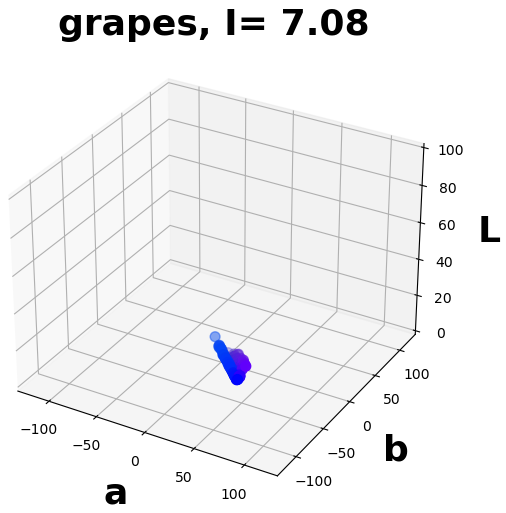}
  \end{subfigure}
  \begin{subfigure}[b]{0.32\linewidth}
    \centering
    \includegraphics[width=\linewidth]{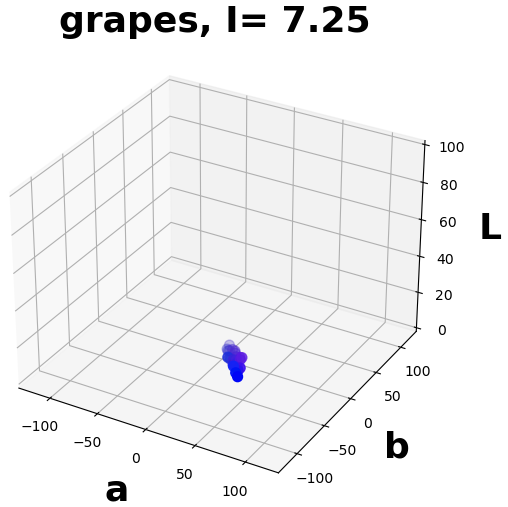}
  \end{subfigure}
    \vspace{1em}

  \begin{subfigure}[b]{0.32\linewidth}
    \centering
    \includegraphics[width=\linewidth]{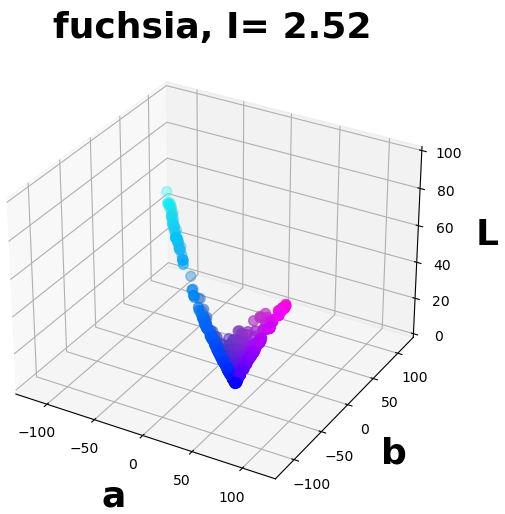}
  \end{subfigure}
  \begin{subfigure}[b]{0.32\linewidth}
    \centering
    \includegraphics[width=\linewidth]{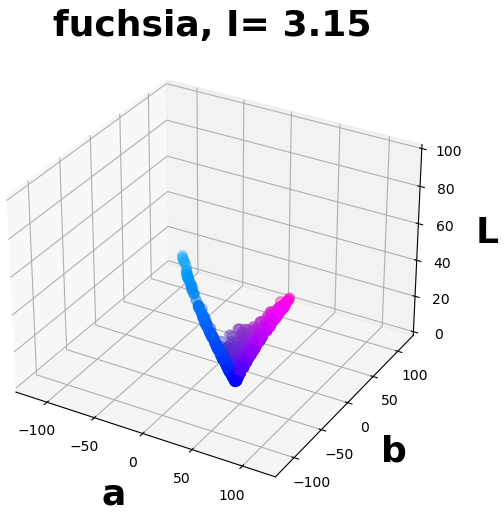}
  \end{subfigure}
  \begin{subfigure}[b]{0.32\linewidth}
    \centering
    \includegraphics[width=\linewidth]{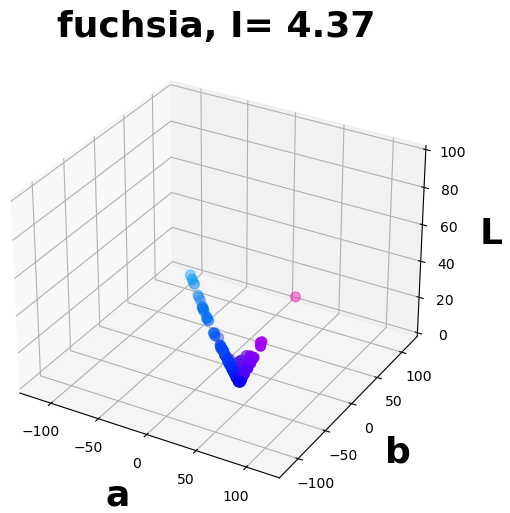}
  \end{subfigure}
    \vspace{1em}

  \begin{subfigure}[b]{0.32\linewidth}
    \centering
    \includegraphics[width=\linewidth]{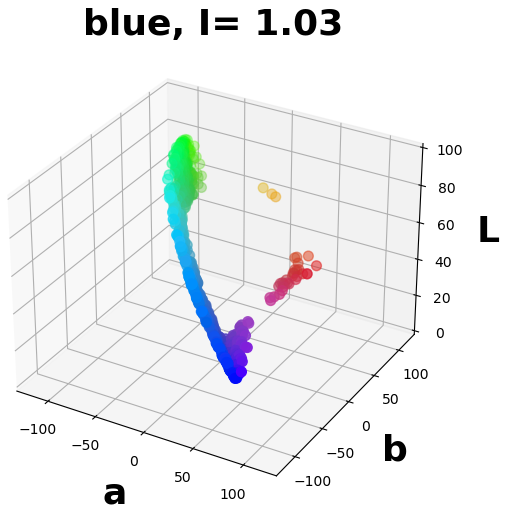}
    \caption{AllFar}
  \end{subfigure}
  \begin{subfigure}[b]{0.32\linewidth}
    \centering
    \includegraphics[width=\linewidth]{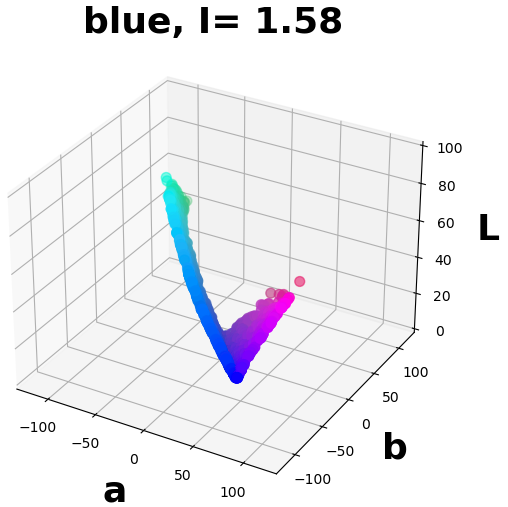}
    \caption{HalfHalf}
  \end{subfigure}
  \begin{subfigure}[b]{0.32\linewidth}
    \centering
    \includegraphics[width=\linewidth]{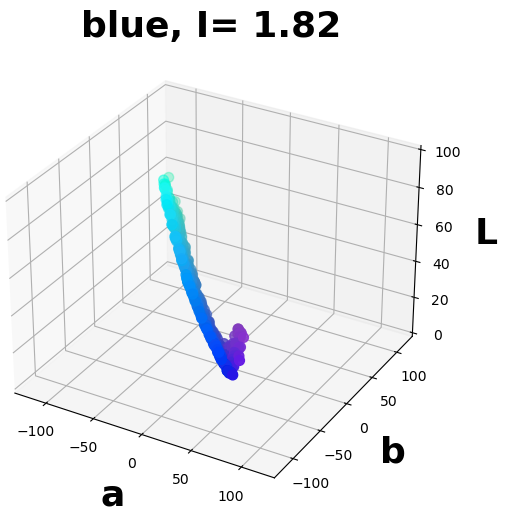}
    \caption{AllClose}
  \end{subfigure}
    \vspace{1em}

  \caption{Denotation in the CIELAB color space for four color terms (\textit{mustard}, \textit{grapes}, \textit{fuchsia}, \textit{blue}) after three RL training conditions.}
  \label{fig:color_examples}
\end{figure}

Figure~\ref{fig:color_examples} illustrates the denotation in the CIELAB color space for four color terms. The visual space covered by the color name varies across RL conditions: in the AllFar condition, each color term covers a broader region, reflecting lower informativeness. In contrast, in the AllClose condition, the denotation is more restricted, indicating higher informativeness.

\label{app:3d_cielab}

\end{document}